\pdfoutput=1
\documentclass[11pt]{article}

\usepackage[margin=1in]{geometry}

\usepackage{amssymb}
\usepackage{amsmath}
\usepackage{latexsym}
\usepackage{amsthm}
\usepackage{booktabs}
\usepackage{enumitem}
\usepackage{graphicx}
\usepackage{color}
\usepackage{algorithm}
\usepackage{algpseudocode}
\usepackage{float}
\usepackage{xcolor}
\usepackage{subcaption}
\usepackage{multirow}

\title{Clustering and Token Denoising for Faster and More Robust VLMs}

\author{
Baptiste Rossigneux$^{1,2}$ \and
Inna Kucher$^1$ \and
Vincent Lorrain$^3$ \and
Emmanuel Casseau$^2$
}

\date{
$^1$Université Paris-Saclay, CEA, LIST, France\\
$^2$Univ. Rennes, CNRS, INRIA, IRISA, France\\
$^3$Vsora, France
}

\begin{document}

\maketitle

\begin{abstract}
Recent Visual-Language Models (VLMs) have enhanced the capabilities of pre-trained LLMs by adding vision tokens alongside text, with approaches like LLaVA showing impressive results. However, the computational burden of processing up to 576 or 729 visual tokens makes edge deployment challenging. While various token pruning techniques require retraining, some are training-free and thus can easily adapt to architecture changes. We introduce ClustRS, a two-part, training-free algorithm for robust token pruning. Its first component is an attention-weighted, clustering algorithm that selects representative tokens from each semantic cluster. The second component, Residual Shrinkage, is a one-pass denoising step on the selected tokens. These training-free lightweight steps make LLaVA ready for real-world data, improving robustness to a wide range of image-noise types and intensities. Experimental results on the ScienceQA-IMG and MM-VET benchmarks show our method outperforms attention- and diversity-based methods by up to 20\% under extreme noise and token conditions (reducing tokens by 97\%, down to 16 tokens) on LLaVA 1.5 7b and achieves exceptional results on LLaVA-OneVision, where we match baseline performance with fewer than one-third of their tokens under mild noise conditions. Our study demonstrates a simple yet powerful alternative to both score-only and diversity-only pruning rules, paving the way for compute-efficient and noise-resilient VLM deployment.
\end{abstract}

\section{Introduction}

The advent of Large Language Models (LLMs) \cite{NIPS2017_3f5ee243, touvron2023llamaopenefficientfoundation} has unlocked powerful reasoning and world knowledge capabilities learned from text. Extending these benefits to vision has led to Visual-Language Models (VLMs), which process both images and text to generate textual outputs. While end-to-end trained models like Flamingo \cite{alayrac2022flamingo} or VILA \cite{liu2024nvila} offer tight integration, they demand extensive computational resources and large multimodal datasets \cite{Dou_2022_CVPR} to adapt. A more resource-efficient paradigm involves modular architectures like LLaVA \cite{NEURIPS2023_6dcf277e, liu2024llavanext, li2024llavaonevisioneasyvisualtask, cocchi2025llavamorecomparativestudyllms}, which leverage pre-trained vision encoders (e.g., CLIP \cite{radford2021learning}, SigLIP \cite{zhai2023sigmoid}) and LLMs connected via lightweight projections. These achieve competitive performance with minimal retraining and easy model swapping.
However, a significant bottleneck persists even in modular VLMs: the sheer number of visual tokens (often 576 or 729) generated by the vision encoder imposes a heavy computational burden on the subsequent LLM, particularly its attention mechanisms. This makes efficient deployment on edge devices challenging. Consequently, various token pruning or compression techniques have emerged. These broadly fall into two categories: training-based approaches, which learn specialized modules but necessitate costly retraining, and more recent training-free methods. The latter offer lightweight alternatives by selecting a subset of tokens post-encoding without further model updates.
Yet their performance under real-world image noise remains under-explored. This raises two questions: (i) how badly does noise degrade current training-free pruning, and (ii) can we design a selector that is both lightweight and resilient to diverse corruptions?
To address these questions, we introduce two lightweight, training-free components. First, Clustering + Huber distance (C + H), an attention-weighted, robust k-means++ procedure that selects diverse yet salient token representatives resilient to outliers caused by noise. Second, Robust Residual Shrinkage (RRS), a refining step on selected tokens by dampening their noise component without shifting their core semantics. For newer SigLIP-based models like LLaVA-OneVision, we further enhance C+H with a salience term to counteract flatter attention distributions (Appendix~A).

Our main contributions are as follows:
\begin{itemize}
  \item Motivated by our analysis of the diversity–quality trade-off in existing token-pruning methods, we introduce a novel robust clustering procedure (C + H) for token selection in LLaVA.
  \item We propose Robust Residual Shrinkage (RRS), a lightweight, one-pass, training-free enhancement that selectively dampens the noise component of each chosen token in feature space, preserving semantic directions and avoiding distributional shift.
  \item We demonstrate that ClustRS, the combination of C\,+\,H and RRS achieves consistent gains in MM-VET under a wide range of Gaussian and Salt and pepper corruptions while incurring negligible extra compute. We show our method outperforms state-of-the-art methods under moderate to heavy noise.
  \item Our entire pipeline being training-free, and preserving full compatibility with existing inference optimizations (e.g.\ KV-cache or flash-attention), it is immediately deployable for efficient, robust VLM inference.
  \item Our method can easily be applied to the newer LLaVA-OneVision model (SigLIP-based), requiring minimal adaptation (see Appendix~A).

\end{itemize}

\section{Related works}

\subsection{Training-based token pruning approaches}

Training-based approaches optimize token selection by learning specialized projector modules during training. Honeybee \cite{Cha_2024_CVPR} introduces "locality-enhanced projectors" that preserve spatial relationships while reducing token count. Using convolution and deformable attention, it maintains local visual context critical for spatial understanding while achieving computational efficiency. TokenPacker \cite{tokenpacker} employs a coarse-to-fine scheme to pack detailed information into compact tokens, first interpolating visual features as low-resolution point queries, then using a region-to-point injection module to infuse high-resolution details from multi-level features. Similarly, TwigVLM \cite{shao2025growingtwigacceleratelarge} "grows" a lightweight trainable block upon an early layer of the base VLM, providing better attention signals for token selection and enabling self-speculative decoding for faster generation. All these approaches significantly outperform naive token selection methods with 75-90\% token pruning, but require additional training components. TokenPacker compresses visual tokens by 75-89\% while maintaining or improving performance with significantly higher efficiency, while TwigVLM achieves 96\% of original performance while pruning 88.9\% of visual tokens and delivers 154\% speedup for long-response generation.

\subsection{Training-free token pruning approaches}

Training-free approaches eliminate redundant visual tokens without requiring additional training. FasterVLM \cite{zhang2024fastervlm} leverages the [CLS] token's (short for "classification token," which serves as an aggregate representation of the image) attention from the visual encoder instead of relying on text-visual attention, which suffers from "attention shift" and "attention dispersion" problems. This approach identifies important visual tokens more accurately by evaluating tokens that receive higher attention from the [CLS] token, achieving 90\% of original performance while pruning 95\% of visual tokens. DivPrune \cite{alvar2025divprune} takes a fundamentally different approach by formulating token pruning as a Max-Min Diversity Problem (MMDP). Rather than prioritizing tokens with high attention scores, it selects tokens that maximize the minimum distance between any pair in the subset, effectively covering the semantic space more comprehensively. This diversity-focused strategy proves particularly effective at high compression ratios, preserving up to 90\% of performance while pruning 80\% of tokens. VisionZip \cite{yang2024visionzip} similarly observes that only a few tokens receive high attention in popular vision encoders and proposes a text-agnostic method to extract more informative tokens. It selects dominant tokens that aggregate substantial information based on attention scores, then merges remaining tokens based on similarity to create contextual tokens, preserving 94\% of performance while pruning 90\% of tokens. Yang et al. \cite{yang2025beyond} challenges intermediate-state-based pruning methods by examining visual tokens' direct impact on the model's output probabilities. Through token- and context-level analyses, these methods detect redundant prototypes, store them in a codebook, and discard visually similar tokens at inference. All four surpass earlier pruning schemes while remaining training-free and computationally light.
A complementary study on Pixel-Value Prediction (PVP) shows that VLMs with unaltered CLIP features still miss pixel-level detail; accuracy rises only after the vision encoder is tuned \cite{gou2024visionlanguagemodelsimage}. This strengthens our claim that post-selection refinement is essential whenever full retraining is not feasible.

\paragraph{Efficient visual representation learning}
Before the recent development of visual-token compression methods for VLMs, several studies investigated token redundancy in ViTs. TokenLearner~\cite{ryoo2021tokenlearner} learns a compact set of input-dependent visual tokens, DynamicViT~\cite{rao2022dynamicvit} progressively prunes low-importance tokens, and ToMe~\cite{bolya2023tome} accelerates pretrained models by merging similar tokens without retraining. STViT~\cite{huang2022stvit} similarly constructs compact semantic tokens from groups of image tokens. These approaches motivate compact visual representations (whether through selection, halting, fusion or merging) by also remaining specific to the ViT architecture. Our method ClustRS performs training-free token selection and denoising for vision-language model inference under image corruption.

\subsection{Works on VLM robustness}
The current research around VLM has focused primarily on defending against adversarial attacks, where malicious perturbations are created to induce specific model failures. A majority of these defense strategies involve a training process, whether it implies a computationally expensive adversarial training or a specialized fine-tuning on a safety-oriented dataset. An example of this paradigmatic focus is the work of Zhang et al. \cite{11072221}, which investigates the robustness of the projector (the multimodal alignment module) itself, instead of focusing on the visual encoder. They introduce an alignment perturbation strategy, which introduces perturbations directly to the intermediate multimodal embeddings. Their method is essentially a fine-tuning they call alignment robust training, where the backbones are frozen and only the lightweight alignment module is trained for better robustness.

Closer to our interest, Wang et al. \cite{wang2025safeguardingvisionlanguagemodelsmitigating} identify that VLMs are surprisingly vulnerable to simple Gaussian noise. In order to address this, they propose a noise-augmented safety fine-tuning on a built dataset called Robust-VLGuard, which includes examples of image-text misalignment. To improve the generalizability of their defensive approach, they use a diffusion model to "purify" adversarial images. Although this approach increases the inference cost, they find that this approach transforms a structured adversarial noise into a less harmful, Gaussian-like noise, which is easier to handle.

In contrast to these training-based defense methods, a representative study by Shirnin et al.~\cite{10526392} conducts a comprehensive robustness analysis by applying a wide range of standard, non-adversarial corruptions to both visual and textual inputs. Their approach is strictly black-box, simulating real-world scenarios where a user has no access to model internals. By evaluating performance degradation on the Visual Question Answering (VQA) task, they identify model-specific weaknesses: for instance, they find that models with separate processing streams for each modality are more susceptible to input noise than single-stream models. Their work provides a valuable diagnostic benchmark, cataloging how standard data corruptions affect different VLM architectures.

Our work builds upon these insights and complements in two ways. First, like Shirnin et al., we focus on robustness to common, non-adversarial noise, but we move beyond a purely diagnostic analysis to propose a novel mitigation strategy. Second, unlike the training-based defenses of Zhang et al. and Wang et al., our proposed method is entirely training-free, designed to be a lightweight enhancement that can be applied at inference time. Previous works do not explore the interplay between computational efficiency and noise resilience. Our work addresses this crucial point by showing that a training-free approach can help for both speed and robustness.

\section{Problem formulation}

\subsection{Token Count and Computational Cost}
\label{subsec:flops-vs-tokens}

The computational cost of Vision-Language Models is primarily driven by the number of visual tokens ($k$) processed by the transformer-based language model. Each token contributes to the quadratic complexity of the self-attention mechanism within transformer architectures, making token count a critical factor in both computation and latency. Consequently, processing a large number of tokens not only demands significant computational resources but also adversely impacts inference speed, particularly on edge or resource-constrained devices where efficiency is paramount.

\vspace{4pt}\noindent\textbf{Practical impact:} Reducing the number of visual tokens significantly improves efficiency. For instance, FasterVLM reports a 4x reduction in storage and approximately a 2.7x inference speedup when reducing token count by 75\%, and a 20x storage reduction alongside a 4x speedup when reducing tokens by 95\%. Minimizing token count is then crucial for efficient inference, motivating our robust token selection and denoising methods for extremely low token numbers.

\subsection{Manifold visualization}

To understand the behavior of token pruning mechanisms, we first visualize the token embeddings in a low-dimensional space and analyze the impact of noise on their distribution. We examine two state-of-the-art methods with contrasting selection philosophies: FasterVLM, which prioritizes attention scores, and DivPrune, which maximizes semantic diversity. This visualization allows us to compare how their distinct strategies translate to the final set of selected tokens, and more important for our study, how these selections are affected by noise corruption. To this end, we use two complementary dimensionality reduction techniques: Principal Component Analysis (PCA) and Uniform Manifold Approximation and Projection (UMAP)~\cite{2018arXivUMAP}. PCA provides a linear projection that preserves the global variance and structure of the data, making it effective for visualizing large-scale distances and the overall token layout. In contrast, UMAP is a non-linear technique that excels at preserving the local neighborhood structure of the data manifold. By presenting both views, we can simultaneously inspect the global token arrangement (via PCA) and the integrity of local semantic clusters (via UMAP), offering a more comprehensive picture of how noise and pruning methods interact.

\begin{figure}
  \centering
  \subcaptionbox{PCA (clean)\label{fig:manifold_pca_clean}\\[1ex]}{%
    \includegraphics[width=0.48\columnwidth]{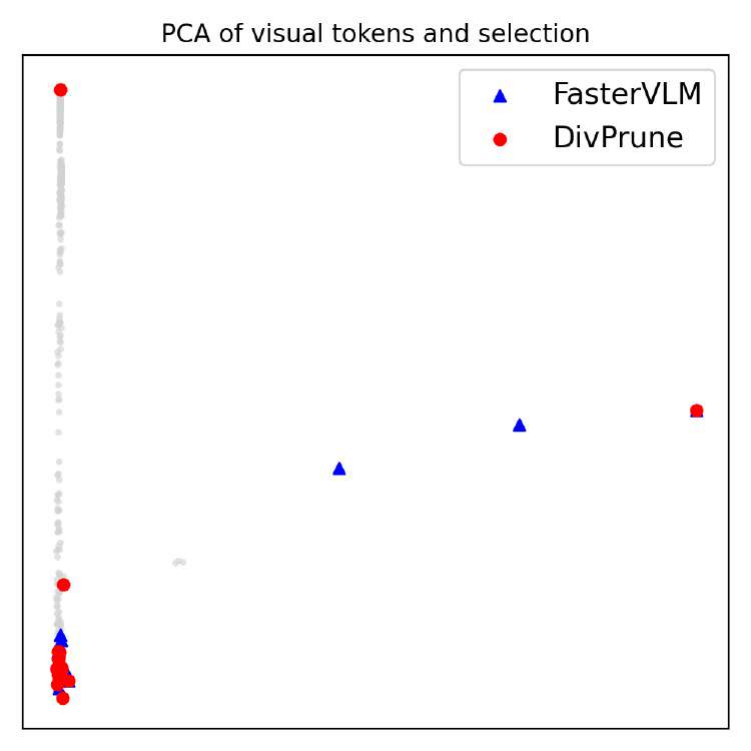}%
  }\hfill
  \subcaptionbox{UMAP (clean)\label{fig:manifold_umap_clean}\\[1ex]}{%
    \includegraphics[width=0.48\columnwidth]{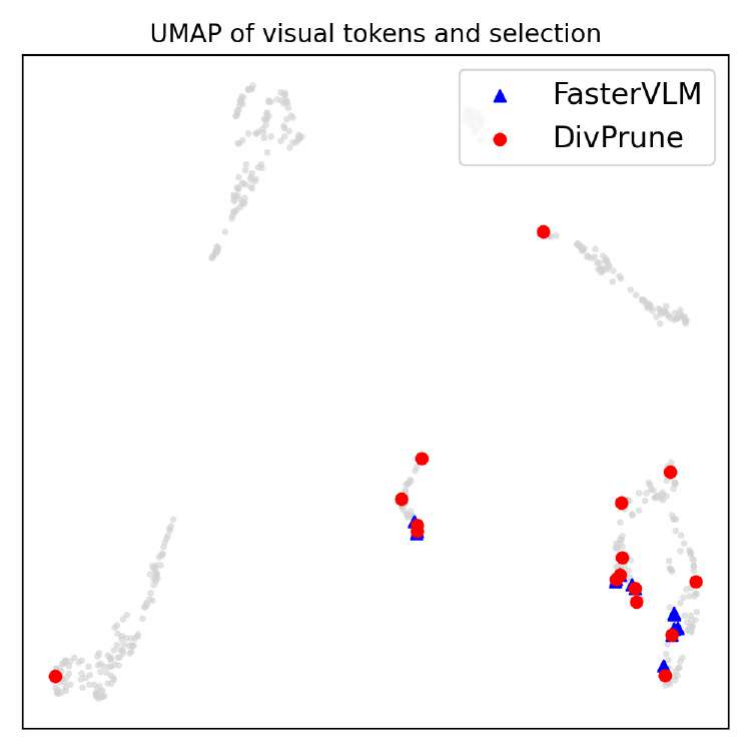}%
  }\\[1ex]
  \subcaptionbox{PCA ($\sigma=3.0$)\label{fig:manifold_pca_noise}\\[2ex]}{%
    \includegraphics[width=0.48\columnwidth]{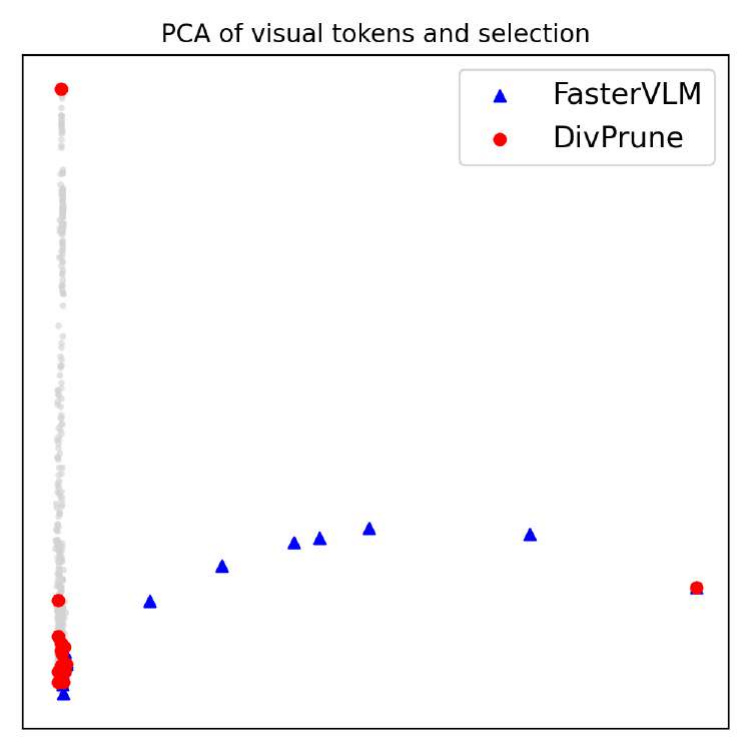}%
  }\hfill
  \subcaptionbox{UMAP ($\sigma=3.0$)\label{fig:manifold_umap_noise}\\[2ex]}{%
    \includegraphics[width=0.48\columnwidth]{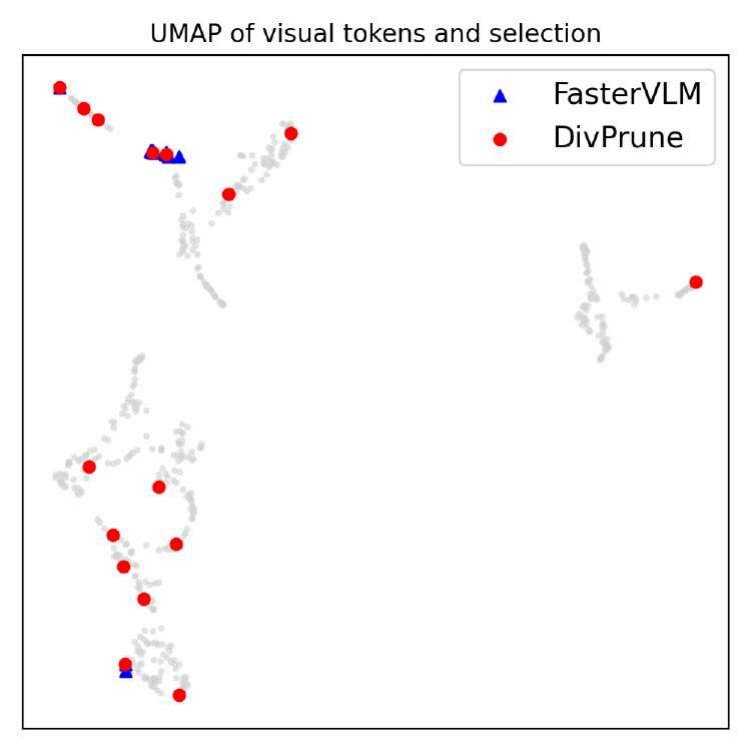}%
  }\\[1ex]
  \caption{%
Two views of the token distribution before and after corruption on the inputs of a LLaVA-1.5-7b.  
\textbf{(a,c) PCA} and \textbf{(b,d) UMAP} plot the same 576 visual tokens in 2-D; gray dots denote all tokens, coloured dots denote those retained by competing pruning rules.  
Top row: clean image. Bottom row: the same image with heavy Gaussian noise ($\sigma\!=\!3.0$).  
Note how noise collapses several semantic groupings into one, illustrating why a purely diversity-based selector can over-allocate tokens under severe corruption.
\\[1ex]
  }
  \label{fig:manifold_comparison}
\end{figure}

Figure~\ref{fig:manifold_comparison} illustrates two key phenomena. First, DivPrune’s red markers lie spread across every semantic “island,” reflecting its pure Max–Min diversity objective. In contrast, FasterVLM (blue) often clusters multiple tokens within the same island, revealing its redundancy-focused criterion. Second, under heavy Gaussian noise ($\sigma=3.0$), the gray islands themselves begin to merge and coalesce—there are simply fewer distinct clusters to cover. In this regime, a diversity-only strategy risks over-allocating budget to a shrinking set of regions, motivating our approach’s need to balance structural coverage with local quality.

\subsection{Why \textsc{FasterVLM} Selects Redundant Tokens}
\label{subsec:redundant-tokens}

\noindent
The \textsc{FasterVLM} pruning rule ranks tokens purely by attention scores with the [CLS] token, averaged over the heads.  
We conduct a controlled corruption study consisting of observing the changes in perceived importance of tokens, ranked by FasterVLM, with the same images and increasing Gaussian noise. This study (Tab.~\ref{tab:rank-change}) reveals that those scores live on an extremely flat landscape: a mean perturbation on the attention values of only $1.6\times 10^{-3}$ (induced by a mild Gaussian noise of $\sigma=0.1$ on the inputs) is sufficient to reshuffle \textbf{87\,\%} of the tokens chosen at a token budget of $k{=}15$.  
Because neighbouring patches often carry near-identical semantics, the displaced tokens are replaced by ``look-alikes,'' so global accuracy hardly moves, masking the numerical brittleness of the top-$k$ sort. As soon as the redundancy is reduced (smaller $k$) or the perturbation grows (e.g.\ $\sigma\!\ge\!1.0$), this silent failure mode surfaces as a sharp performance drop, motivating the need for ranking-stable, structure-aware selection.

\begin{table}[!htbp]
    \centering
    \small
    \setlength{\tabcolsep}{8pt}
    \begin{tabular}{lcc}
        \toprule
        Noise level $\sigma$ & Avg.\ rank change (top–15) & Top–15 overlap \\
        \midrule
        $0.0$ & $0$   & $100\,\%$ \\
        $0.1$ & $288$ & $13\,\%$  \\
        $1.0$ & $321$ & $0\,\%$   \\
        $3.0$ & $308$ & $0\,\%$   \\
        \bottomrule
    \end{tabular}
    \caption{Instability of FasterVLM's token ranking under increasing Gaussian noise.
    \textbf{Avg. rank change (top-15)}: The average absolute change in rank of importance for the original top-15 tokens after noise is applied. For example, a token ranked 1st by FasterVLM on a clean image could drop to rank 300 even with a small noise.
    \textbf{Top-15 overlap}: The percentage of tokens that remain in the top-15 set after noise is applied.}
    \label{tab:rank-change}
\end{table}

\medskip
\noindent
Pure ``quality'' scoring (with attention) without an explicit coverage term yields brittle selections; conversely, methods that look for coverage alone (e.g.\ \textsc{DivPrune}) ignore token salience and often keep uninformative patches. DivPrune solves a pure Max–Min objective and therefore excels at geometric coverage, but its diversity-only criterion is oblivious to token salience; in low-budget or low-noise settings this leads to trading high-attention evidence for distant yet uninformative patches, a complementary failure mode to FasterVLM’s redundancy.

\section{Robust Token Selection Methodology}

\begin{figure}
    \centering
    \begin{minipage}{0.48\textwidth}
        \centering
        \includegraphics[width=\textwidth]{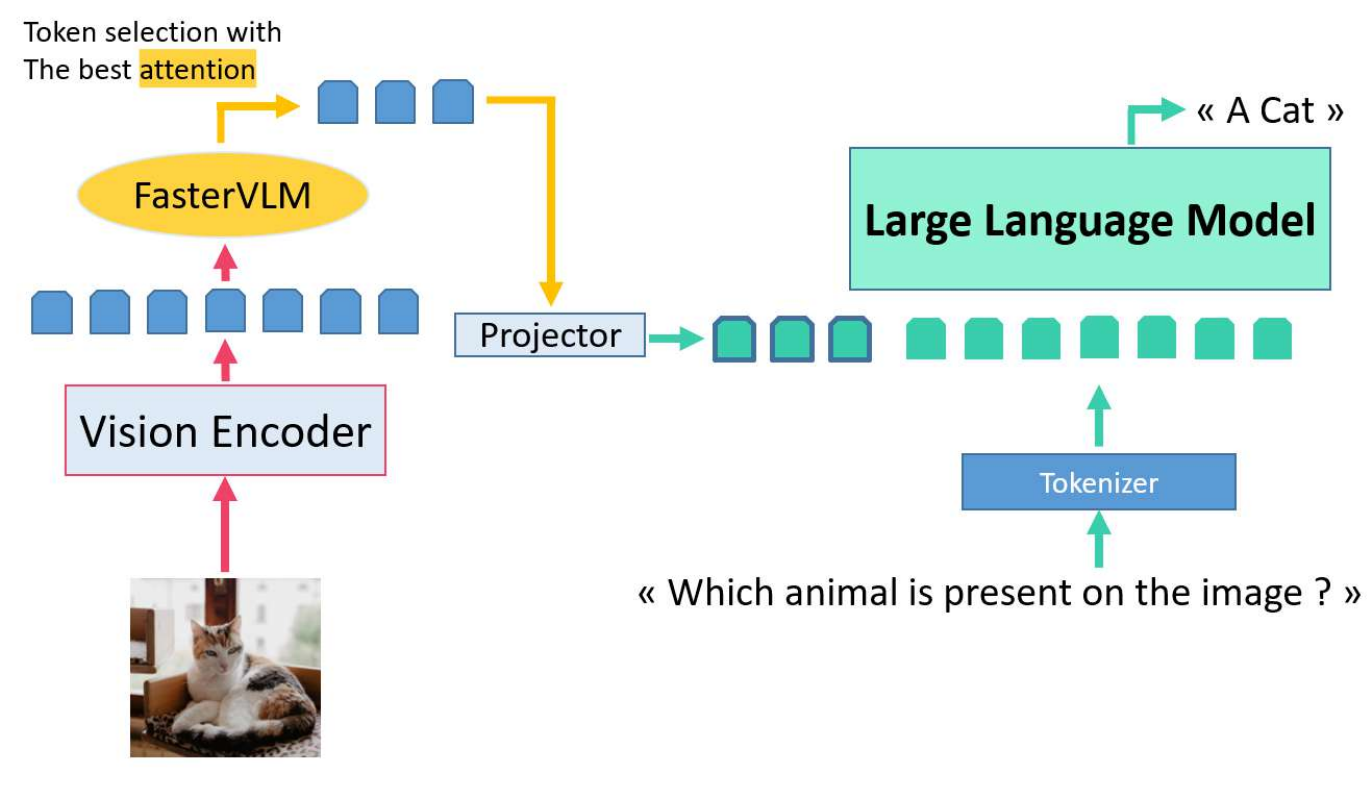}
    \end{minipage}
    \hfill
    \begin{minipage}{0.02\textwidth}
        \centering
        \vspace{0.2cm} 
        \color[gray]{0.7}\rule{0.5pt}{4cm}
    \end{minipage}
    \hfill
    \begin{minipage}{0.48\textwidth}
        \centering
        \includegraphics[width=\textwidth]{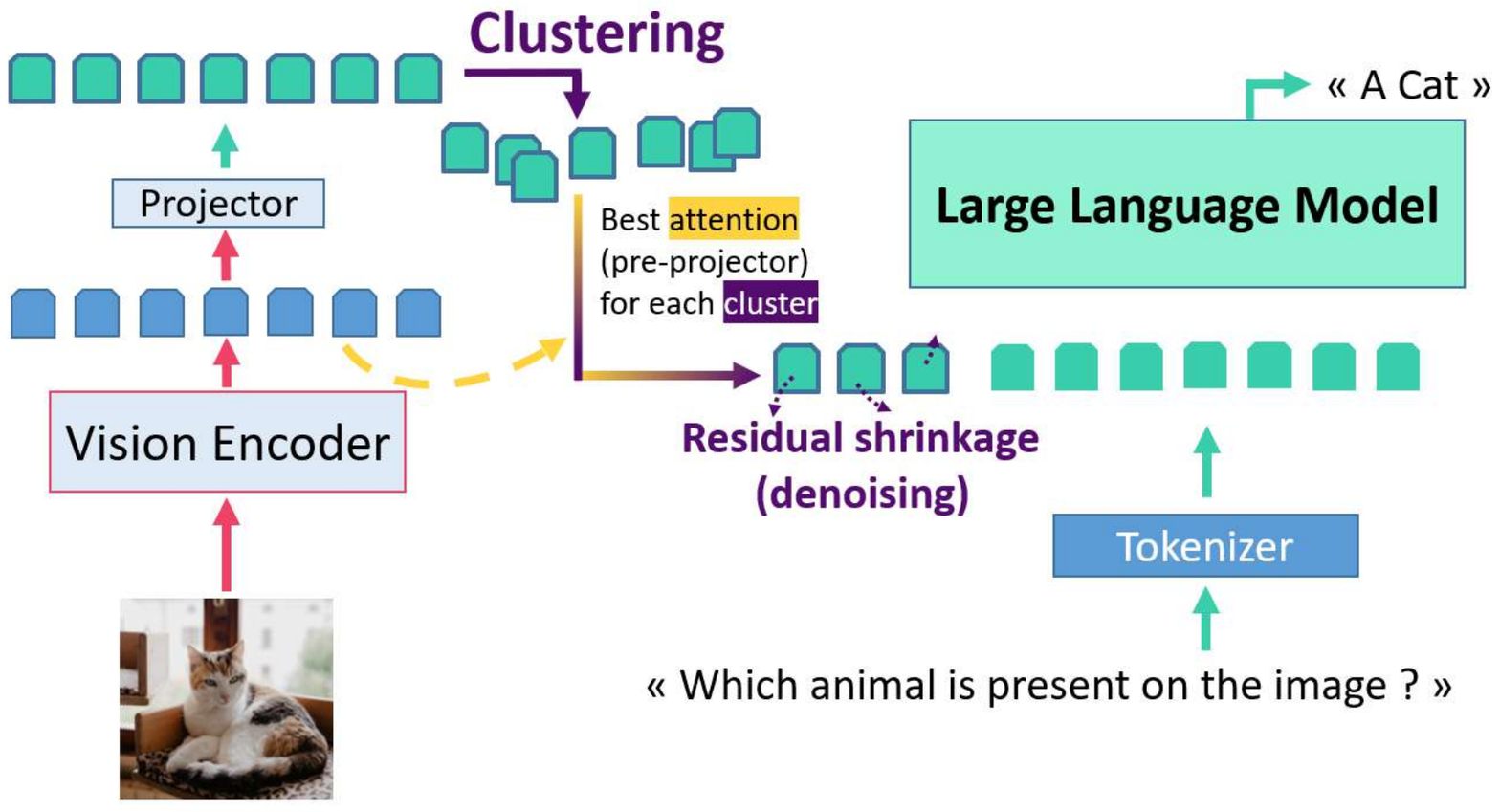}
    \end{minipage}
    \caption{High-level comparison of the token–selection pipelines. 
\textbf{Left:} FasterVLM keeps the $k$ tokens with the highest \texttt{[CLS]} self-attention scores immediately after the vision encoder. 
\textbf{Right:} our C\,+\,H module first applies attention-weighted Huber $k$-means to cluster the projected tokens and then retains a single, most important representative from each cluster, with the help of the \texttt{[CLS]} self-attention scores from the pre-projection representations. This reduced set is then enhanced with RRS to denoise each token with its cluster statistics, before being forwarded to the language model.}
    \label{fig:both_images}
\end{figure}

\subsection{Clustering Tokens with Attention and Huber Distance (C\,+\,H)}
\label{sec:ch}

A single $336{\times}336$ image processed by CLIP becomes $N = 576$ visual tokens (from a grid of $24{\times}24$ patches, where each patch is $14{\times}14$ pixels). Because the language‐model decoder incurs $\mathcal{O}(k^{2})$ FLOPs once these tokens enter the self-attention stack, we must cut $k$ down sharply at inference time. Our answer is to cluster the tokens and keep one representative per cluster, enforcing semantic diversity: each survivor stands for a distinct region of feature space.

I perform this clustering after the vision-language projection layer, a choice that is fundamental to my method's design (see Figure \ref{fig:both_images}). The projection layer is specifically trained to translate visual features from the CLIP encoder's space into the semantic embedding space of the language model. This means that distances and densities in the post-projection space reflect the LLM's own understanding of conceptual relationships. Clustering in this space allows us to group tokens based on the very semantic similarities the LLM will use for reasoning. In contrast, clustering on raw CLIP features would group tokens based on visual-contrastive similarity, which does not always correlate with the semantic nuances required for complex textual question-answering. My approach therefore ensures that the token selection process is directly guided by the patterns of the downstream language model.

The same clustering and selection procedure generalizes to SigLIP-based backbones with a minor adaptation described in Appendix~A.

\subsubsection{Attention-Weighted Clustering Objective}

Not all visual tokens are of equal value. The vision encoder provides a crucial hint by assigning each token~$\mathbf{v}_i$ a scalar self-attention score $a_i \ge 0$ that correlates with its importance. Our goal is to leverage this information to guide the clustering process. We achieve this by modifying the standard k-means objective to an attention-weighted within-cluster distortion, which we aim to minimize:
\begin{equation} \label{eq:attn-wkmeans}
\mathcal{L} = \sum_{j=1}^{k} \sum_{i : c(i)=j} a_i \, d(\mathbf{v}_i, \mathbf{c}_j)^2
\end{equation}
Here, the components are defined as follows:
\begin{itemize}
    \item The outer sum $\sum_{j=1}^{k}$ iterates over each of the $k$ clusters.
    \item The inner sum $\sum_{i : c(i)=j}$ iterates over all visual tokens $\mathbf{v}_i$ that have been assigned to the $j$-th cluster.
    \item $a_i$ is the pre-computed attention score of token~$\mathbf{v}_i$, acting as its importance weight.
    \item $d(\mathbf{v}_i, \mathbf{c}_j)^2$ is the squared Euclidean distance between a token~$\mathbf{v}_i$ and the centroid~$\mathbf{c}_j$ of the cluster it belongs to.
\end{itemize}
In essence, tokens with a higher attention score $a_i$ contribute more significantly to the total loss, and therefore exert a stronger pull on their cluster's centroid during optimization.

\label{sec:ch-baseline}

\subsubsection{K-means++ for Immediate Diversity}

Vanilla $k$-means is notoriously sensitive to initialisation; poor seeds can trap the algorithm in redundant, high-loss states. We therefore adopt the $k$-means++ strategy, modified to be attention-aware: the first centroid is the single highest-attention token, and each subsequent centroid is sampled with a probability $p_i \propto a_i \cdot d(\mathbf{v}_i, \mathbf{C})^2$, where $\mathbf{C}$ is the set of already chosen centroids. This simple adjustment spreads the seeds across feature space in proportion to both distance and attention, giving the refinement phase a good start toward well-separated, information-rich clusters. After seeding, we run an iterative refinement loop that alternates between two steps:
\begin{itemize}
    \item \textbf{Assignment:} Each token $\mathbf{v}_i$ is assigned to the cluster of its nearest centroid $\mathbf{c}_j$, based on Euclidean distance.
    \item \textbf{Update:} Each centroid $\mathbf{c}_j$ is re-calculated as the attention-weighted average of all tokens assigned to it.
\end{itemize}
This loop runs for a small, fixed number of iterations ($I \le 3$ in our experiments) or until fewer than $1\%$ of tokens switch clusters. The effectiveness of the k-means++ initialization is what allows for allows for this minimal number of refinement steps: often just one or two to achieve high-quality clusters.

So far we have a diversity-aware selector, usable to discard every tokens but the ones with highest attention for each clusters. However, this selector it is still vulnerable to outlier tokens introduced by image corruptions.

\subsubsection{Adding Huber distance for outlier-robust clusters}

Image corruptions (namely Gaussian blur and sensor noise) push a handful of tokens far away from their clean locations in feature space. With the squared‐error objective of Eq.~\eqref{eq:attn-wkmeans} those outliers dominate $\mathcal L$, dragging centroids off the true semantic “islands” (Fig.~\ref{fig:manifold_comparison}).

\paragraph{Robustness to Outliers via the Huber distance}
Under a standard squared-error objective, outliers induced by noise disproportionately influence the loss function, which can skew the cluster centroids away from the dense, semantically meaningful regions of the data manifold (as illustrated in Fig.~1).

To mitigate this vulnerability, we replace the squared Euclidean distance with the Huber distance, a loss function specifically designed for robustness to outliers. The Huber distance $d_{H}$ is defined as:
\begin{equation}
h_\tau(r) = 
\begin{cases} 
      \frac{1}{2}r^2 & r \le \tau \\
      \tau(r - \frac{1}{2}\tau) & r > \tau 
   \end{cases},
\quad \text{where } d_H(\mathbf{v}_i, \mathbf{c}_j) = \sqrt{h_\tau(\|\mathbf{v}_i - \mathbf{c}_j\|_2)}.
\end{equation}
Here, $r$ is the residual magnitude $\|\mathbf{v}_i-\mathbf{c}_j\|_2$, and $\tau$ is the Huber threshold.
We use a shared value of $\tau=1$ across all experiments. A sensitivity analysis over different threshold values is provided in Appendix~\ref{sec:sensitivity}, Table~\ref{tab:tau_sensitivity}.

The Huber distance combines the desirable properties of both squared-error and absolute-error losses. For residuals below the threshold $\tau$, it behaves quadratically, retaining the strong convergence properties of squared loss for inliers and encouraging tight cluster formation. For residuals above $\tau$, the cost transitions to a linear penalty. This transition is critical: it ensures that the influence of large-error outliers is substantially reduced, preventing them from exerting excessive pull on the centroid positions. By incorporating the Huber cost into our clustering objective (replacing $d$ by $d_{H}$), we make the process more resilient to noise-induced artifacts without requiring any changes to the clustering algorithm itself. The improvement over selection with k-means++ using Euclidean distance was measured and reported in our ablation study in Table \ref{tab:ablation_c_h_rrs_noisy}.

\noindent\textbf{Why not cosine?}
Cosine distance ignores vector norms, yet CLIP-style embeddings encode
salience partly in their magnitude.  Huber preserves both angle and
length while still damping large-norm noise.

\subsection{Robust Token Residual Shrinkage (RRS)}
\label{sec:rrs}

\begin{figure}[t]
  \centering
  \subcaptionbox{Global PCA token movement\label{fig:global_move}\\[1ex]}{%
    \includegraphics[width=0.50\columnwidth]{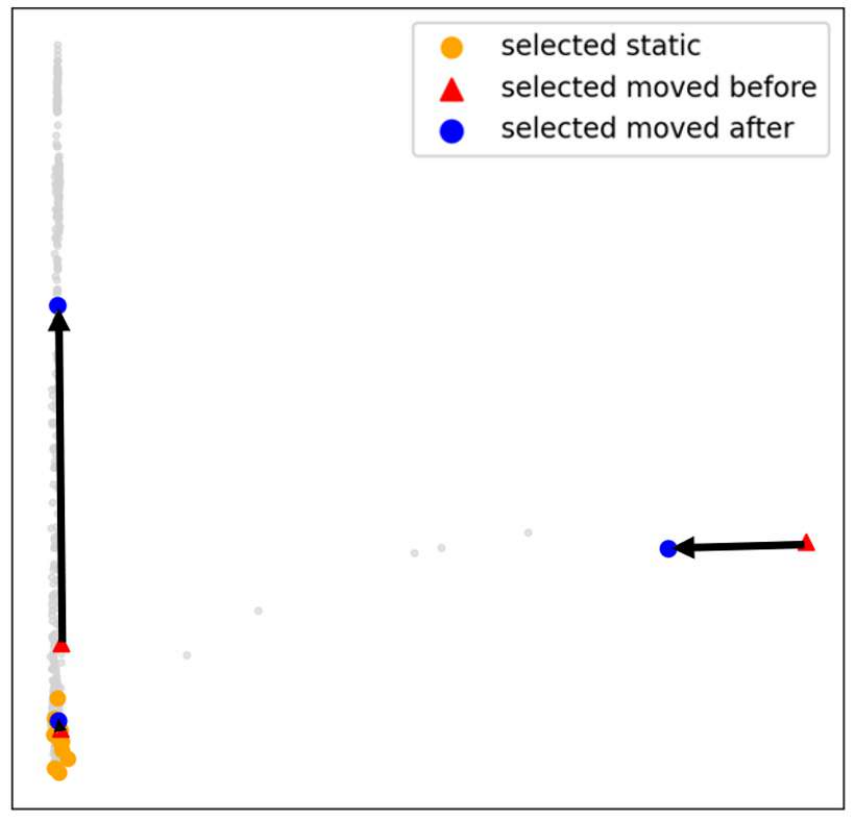}%
  }\hfill
  \subcaptionbox{Cluster‐only PCA movement\label{fig:cluster_move}\\[1ex]}{%
    \includegraphics[width=0.50\columnwidth]{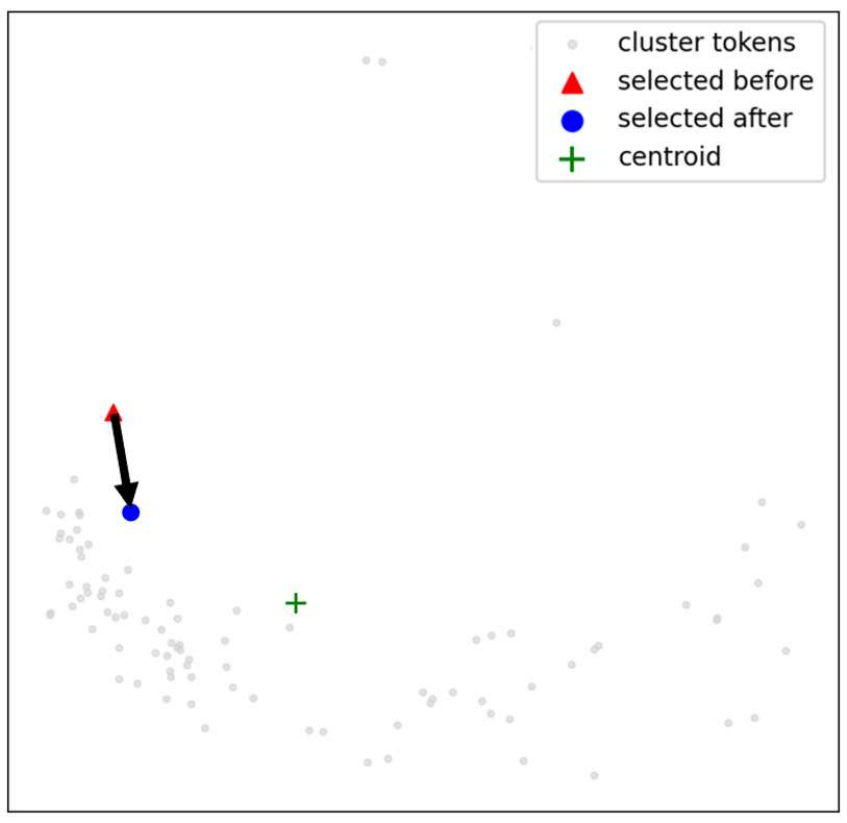}%
  }\\[2ex]
  \caption{%
    Comparison of full‐image vs.\ cluster‐focused token movement after RRS.
    (\textbf{a}) Global PCA on all $N$ tokens highlights which selected tokens have moved (red $\rightarrow$ blue) in context.
    (\textbf{b}) Local PCA fit on one cluster’s tokens shows fine‐grained per‐cluster shifts.\\[1ex]
  }
  \label{fig:movement_comparison}
\end{figure}

Selection alone cannot fix the fact that the remaining tokens are still contaminated by noise. The next step is therefore to denoise those tokens using information from the whole cluster. After robust Huber $k$-means++, each cluster centroid $\mathbf{c}_{j}\in\mathbb{R}^{D}$ is a robust summary of its member tokens.  A selected token vector $\mathbf{v}_{i}$ can therefore be written as the sum of the centroid plus a \emph{residual} $\mathbf{r}_{i}=\mathbf{v}_{i}-\mathbf{c}_{j(i)}$. Under additive input noise, most of the variance lies in this residual. Shrinking an estimate toward a low-variance anchor can lower its mean-squared error whenever the dimension \(D\!\ge\!3\).
Put simply, we will refine every selected token a little closer to the average token in its cluster: enough to smooth out noise, not enough to change its meaning.
Applying this idea in token space gives us a training-free denoising step that preserves each token’s original feature direction while reducing variance (the movement in token-space is illustrated in Figure \ref{fig:movement_comparison}).

\vspace{4pt}\noindent\textbf{Adaptive shrink factor.}\;
For each cluster $j$ we estimate a robust scale
\begin{equation}
  \delta_{j}
  = \operatorname{median}_{k\,:\,c(k)=j}
    \bigl\|\,\mathbf{v}_{k}-\mathbf{c}_{j}\bigr\|_{2},
  \label{eq:delta}
\end{equation}
where $\|\!\cdot\!\|_{2}$ is the Euclidean norm.
Given $\delta_{j}$, the per-token shrink coefficient is
\begin{equation}
  \alpha_{i}
  =\frac{\delta_{j}^{2}}
         {\bigl\|\mathbf{r}_{i}\bigr\|_{2}^{2}\;+\;\delta_{j}^{2}},
  \qquad i\in\mathcal{C}_{j}.
  \label{eq:alpha}
\end{equation}

Tokens already close to the centroid
($\|\mathbf{r}_{i}\|\!\ll\!\delta_{j}$) keep $\alpha_{i}\!\approx\!1$,
while distant (likely noisy) tokens move closer
($\alpha_{i}\!\rightarrow\!0$). This coefficient is clipped to $[0.1, 0.95]$ to avoid collapsing tokens onto the centroid.

\noindent\textbf{Enhanced token.}\;
The denoised representation is a convex combination
\begin{equation}
  \tilde{\mathbf{v}}_{i}
  \;=\;
  \alpha_{i}\,\mathbf{v}_{i}
  \;+\;
  \bigl(1-\alpha_{i}\bigr)\,\mathbf{c}_{j(i)}.
  \label{eq:rrs}
\end{equation}
Because we never re-normalise $\tilde{\mathbf{v}}_{i}$, the
directional statistics expected by the language model remain intact.

\begin{algorithm}
\small
\caption{Residual Token Shrinkage (per image)}
\label{alg:rrs}
\begin{algorithmic}[1]
\Require projected tokens $\{\mathbf{v}_{i}\}_{i=1}^{N}$,
         attention scores $\{a_{i}\}$,
         cluster labels $\{c(i)\}$, token subset $\mathcal{S}$
\ForAll{clusters $j=1,\dots,k$}
  \State $\mathbf{c}_{j}\gets
         \dfrac{\sum_{i:c(i)=j} a_{i}\,\mathbf{v}_{i}}
              {\sum_{i:c(i)=j} a_{i}}$  \Comment{attention-weighted centroid}
  \State $\delta_{j}\gets \operatorname{median}_{i:c(i)=j}
                          \|\,\mathbf{v}_{i}-\mathbf{c}_{j}\|_{2}$
  \ForAll{$i\in\mathcal{S}\;\land\;c(i)=j$}
     \State $\alpha_{i}\gets
            \delta_{j}^{2}\big/\bigl(\|\mathbf{v}_{i}-\mathbf{c}_{j}\|_{2}^{2}
                                      +\delta_{j}^{2}\bigr)$
     \State \textbf{clip} $\alpha_{i}$ to $[0.1,0.95]$
     \State $\tilde{\mathbf{v}}_{i}\gets
            \alpha_{i}\mathbf{v}_{i}+(1-\alpha_{i})\mathbf{c}_{j}$
  \EndFor
\EndFor
\end{algorithmic}
\end{algorithm}

\noindent
RRS adds \(\mathcal{O}(kD)\) floating-point operations, the cost of a few
dot‐products per selected token, negligible compared to the vision tower
forward pass.

\paragraph{Connection to James--Stein estimator}
Our \textbf{Robust Residual Shrinkage (RRS)} method is inspired by the principles of the James-Stein estimator~\cite{james_estimation_1961}, a foundational result in statistical estimation theory. The estimator addresses the problem of estimating multiple Gaussian means $\boldsymbol{\theta} = (\theta_1, \dots, \theta_D)$ from a set of observations $\mathbf{x} = (x_1, \dots, x_D)$. Counter-intuitively, the standard Maximum Likelihood Estimator (MLE), $\hat{\boldsymbol{\theta}}_{\text{MLE}} = \mathbf{x}$, is \textit{inadmissible} for dimensions $D \ge 3$. This means a universally better estimator exists, one that is guaranteed to have a lower total mean squared error (MSE). This superior performance is achieved by \textit{shrinking} the individual estimates towards a common center. 

\textbf{The estimator's logic:} The classical James-Stein estimator improves upon the MLE by shrinking the individual estimates towards a central point (typically the origin). The estimator takes the form:
\begin{equation} \label{eq:james_stein}
\hat{\boldsymbol{\theta}}_{\text{JS}} = \left(1 - \frac{(D-2)\sigma^2}{\|\mathbf{x}\|^2}\right) \mathbf{x}
\end{equation}
The term in parentheses is the \textit{shrinkage factor}. By borrowing strength across all dimensions to inform each individual estimate, the estimator reduces the total error. The risk (expected MSE) of $\hat{\boldsymbol{\theta}}_{\text{JS}}$ is guaranteed to be lower than the risk of $\hat{\boldsymbol{\theta}}_{\text{MLE}}$~\cite{tsybakov2009nonparametric}.

\textbf{Mapping RRS to the James-Stein Framework:} RRS follows this general intuition but is not an application of the classical James--Stein estimator. In particular, we do not assume that visual-token residuals are independent, isotropic, or Gaussian. The noisy token embedding $\mathbf{v}_i$ that we observe corresponds to the statistical observation $\mathbf{x}$, while the ``true,'' noise-free embedding $\mathbf{v}_i^*$ that we aim to estimate represents the mean $\boldsymbol{\theta}$. A crucial adaptation in our method is the choice of the shrinkage center. Instead of shrinking towards the origin, we shrink towards a more informative, low-variance anchor: the cluster centroid $\mathbf{c}_j$. The centroid serves as a robust, data-driven estimate of the semantic concept represented by the cluster. Similarly, our adaptive shrink factor from Eq.~(\ref{eq:alpha}), $\alpha_i = \delta_j^2 / (\|\mathbf{r}_i\|^2 + \delta_j^2)$, plays the same role as the classical shrinkage term, dynamically adjusting the degree of shrinkage based on the token's distance from the centroid.

Consequently, our enhanced token update rule from Eq.~(\ref{eq:rrs}), $\tilde{\mathbf{v}}_i = \alpha_i \mathbf{v}_i + (1 - \alpha_i) \mathbf{c}_j$, can be rewritten as $\tilde{\mathbf{v}}_i = \mathbf{c}_j + \alpha_i (\mathbf{v}_i - \mathbf{c}_j)$. This reformulation makes it clear that we are shrinking the residual vector $\mathbf{r}_i = \mathbf{v}_i - \mathbf{c}_j$ by a factor of $\alpha_i$. 

We thus interpret the James--Stein result as motivation for the use of shrinkage, rather than as a theoretical guarantee for RRS. The usefulness of the proposed rule for visual-token denoising is established empirically through the component and parameter ablations reported in Table~\ref{tab:ablation_c_h_rrs_noisy}.

\section{Experiments}

\subsection{Experimental setup}

All experiments reported here were conducted using LLaVA-1.5-7B (CLIP-based). Complementary results demonstrating the effectiveness of our approach on LLaVA-OneVision (SigLIP-based backbone) are available in Appendix~A.

\subsubsection{Datasets}
We evaluate our method on two challenging visual reasoning benchmarks: MM-VET and ScienceQA-IMG.

\paragraph{MM-VET} The Multimodal Visual Evaluation Task (MM-VET) \cite{yu2024mm} is designed to assess the open-ended visual reasoning capabilities of VLMs. In MM-VET, models receive image-based questions requiring descriptive, explanatory, or reasoning-based answers. Unlike constrained tasks, MM-VET permits free-form responses, with an independent LLM used as a judge to evaluate correctness. This setup mimics a real-world scenario where models must generate coherent, contextually appropriate answers without predefined options, making it inherently more difficult. With MM-VET, the judge is traditionally GPT-4. However, because GPT-4 inference is very costly and that new models have since come out as much more performant and cheaper, we decided to instead use Deepseek-V4 Pro~\cite{deepseekv4technicalreport}. This model is much more performant on every benchmark, and orders of magnitude cheaper than GPT-4.

\paragraph{ScienceQA-IMG} 
The ScienceQA-IMG dataset is derived from the broader ScienceQA dataset~\cite{lu2022learn} and contains only questions accompanied by an image and poses them in a multiple-choice format. Because answer candidates are pre-specified, many items can already be answered with linguistic priors (or partial text clues) alone; indeed, prior work reports respectable accuracy even when the image is withheld. Consequently, the dataset is far less sensitive to fine-grained visual cues than MM-VET, and performance margins between pruning methods tend to be smaller. We evaluate models by straightforward top-1 accuracy on the provided choices, following standard practice.

\subsubsection{Noise Simulation}
\label{sec:noise-sim}

To probe robustness we corrupt every test image on–the–fly with a parametric noise model before it enters the vision encoder; the exact same corrupted frame is then fed to all competing methods to ensure fair comparison.

We focus on two canonical families that together cover both additive and impulsive artefacts frequently encountered in real-world camera pipelines:

\begin{enumerate}[leftmargin=1.2em, itemsep=1pt]
\item \textbf{Gaussian} (additive white) noise  
      \[
        I^\text{noisy}(x,y)
          = I(x,y) + \mathcal N\!\bigl(0,\sigma^{2}\bigr),
      \]
      applied independently to each colour channel and pixel, then
      clipped to \([0,255]\).  We evaluate for  
      $\sigma\!\in\!\{0.1,\,0.5,\,1.0,\,2.0,\,3.0\}$,
      expressed in the normalised range \([0,1]\) so
      $\sigma{=}1$ already produces clearly visible grain.

\item \textbf{Salt–\&–pepper} (impulsive) noise  
      Each pixel is independently replaced by black $(0)$ or
      white $(255)$ with probability $\rho/2$ each:
      \[
        I^\text{noisy}(x,y)=
        \begin{cases}
          0            & \text{w.p. } \rho/2,\\
          255          & \text{w.p. } \rho/2,\\
          I(x,y)       & \text{otherwise}.
        \end{cases}
      \]
      We test five densities
      $\rho\!\in\!\{0.01,\,0.1,\,0.2,\,0.3,\,0.5\}$,
      spanning barely perceptible artefacts (1\% pixels flipped) to
      heavy corruption (50 \%).
\end{enumerate}

Additionally, to extend the evaluation beyond additive and impulsive noise, we additionally consider two photometric transformations and one occlusion-based corruption on MM-VET:

\begin{itemize}[leftmargin=1.2em, itemsep=1pt]
\item \textbf{Brightness shift:} image brightness is modified using factors
$b\in{0.5,1.5,2.0}$, covering both darkened and overexposed inputs.

\item \textbf{Contrast shift:} image contrast is modified using factors
$c\in\{0.5,1.5,2.0\}$, representing reduced- and increased-contrast conditions.

\item \textbf{Cutout:} a randomly positioned region is masked, with severity
$p\in\{0.1,0.25,0.4\}$ denoting the fraction of the image area removed.
This corruption approximates partial occlusion or missing visual information.
\end{itemize}

Much like the main corruption experiments, the same corrupted image is provided to each of the methods for fair comparison. These transformations remain controlled corruptions rather than a comprehensive simulation of all real-world distribution shifts, but they still showcase that gains extend to more natural noise sources.

\subsubsection{Baselines and Comparison}
Our method, ClustRS, is benchmarked against FasterVLM, DivPrune and VisionZip. We compare performance across three distinct token budgets ($k \in {16, 50, 144}$) to assess effectiveness under varying computational constraints. We obtained their accuracies by adapting their available code to our setup for fair comparisons.

\subsubsection{Metrics}
Performance on MM-VET is reported using the LLM judge’s accuracy scores, reflecting open-ended visual reasoning capabilities. ScienceQA-IMG performance is quantified using standard multiple-choice accuracy, highlighting model correctness in structured scientific reasoning contexts.

Below we condense the full noise sweep, from small noise to intense noise, into three budget regimes, highlighting how our two contributions, better selection (C\,+\,H) and enhancement (RRS), shift the accuracy landscape.

\subsection{Robustness on MM-VET with LLaVA-1.5-7B}
\label{sec:mmvet-results}

\paragraph{Extreme Compression Regime ($k=16$)}
At this highly constrained budget, corresponding to a 97\% reduction from 576 visual tokens, ClustRS achieves the strongest overall performance. On clean images, it obtains a score of 0.271, outperforming FasterVLM, DivPrune, and VisionZip by 6.0, 3.9, and 3.2 percentage points, respectively. This advantage is largely maintained as the corruption severity increases. Under severe Gaussian noise at $\sigma=3.0$, ClustRS reaches 0.140, compared with 0.132 for FasterVLM and DivPrune and 0.133 for VisionZip. Under heavy salt-and-pepper corruption at $\rho=0.5$, ClustRS scores 0.215, exceeding the second-best method, FasterVLM, by 1.4 percentage points. ClustRS also obtains the highest mean scores under brightness and contrast transformations (0.250) and cutout corruption (0.214). These results indicate that the benefits of combining robust clustering and residual shrinkage are most pronounced when only a very small number of visual tokens can be retained. These conclusions are further showcased on Table~\ref{tab:corruption_performance} which compiles extreme compression results on motion blur and JPEG artifacts, with the same results. 

\paragraph{Moderate Compression Regime ($k=50$)}
With approximately 91\% of the original visual tokens removed, ClustRS provides a favorable balance between clean-image performance and corruption robustness. On clean images, it achieves the highest score of 0.304, improving over FasterVLM, VisionZip, and DivPrune by 1.4, 1.9, and 2.0 percentage points, respectively. Its performance under Gaussian noise is more variable: VisionZip performs better at $\sigma=1.0$, and the two methods are tied at $\sigma=2.0$. Nevertheless, at the most severe Gaussian corruption level, $\sigma=3.0$, ClustRS obtains the highest score of 0.163, narrowly exceeding DivPrune at 0.160. ClustRS is more consistently effective under salt-and-pepper noise, achieving the highest score at every evaluated density. At $\rho=0.5$, it reaches 0.229, compared with 0.221 for FasterVLM, 0.210 for VisionZip, and 0.190 for DivPrune. It also achieves the best results under brightness and contrast transformations (0.292) and cutout corruption (0.233).

\paragraph{Conservative Compression Regime ($k=144$)}
At the more conservative budget of 144 tokens, corresponding to a 75\% token reduction, the differences between methods become smaller and depend more strongly on the corruption type. On clean images, ClustRS obtains 0.320, slightly below FasterVLM and VisionZip, which both achieve 0.322, and DivPrune at 0.318. The Gaussian-noise results are mixed: ClustRS achieves the highest scores at $\sigma=0.1$ and $\sigma=0.5$, but is very slightly outperformed by competing methods at the higher corruption levels. At $\sigma=3.0$, it reaches 0.155, compared with 0.158 for FasterVLM and DivPrune and 0.154 for VisionZip. In contrast, ClustRS consistently obtains the highest score at every evaluated salt-and-pepper density. At $\rho=0.5$, it achieves 0.237, exceeding DivPrune, VisionZip, and FasterVLM by 1.1, 2.3, and 2.9 percentage points, respectively. Under the additional corruptions, ClustRS also obtains the best cutout result (0.267), while its brightness-and-contrast score of 0.305 remains close to the best result of 0.310 obtained by FasterVLM. Overall, these findings suggest that ClustRS remains competitive at larger token budgets, although its advantage is less uniform than in the extreme-compression regime.

\begin{figure}[!htbp]
  \centering
  \begin{minipage}{0.48\linewidth}
    \centering
    \includegraphics[width=\linewidth]{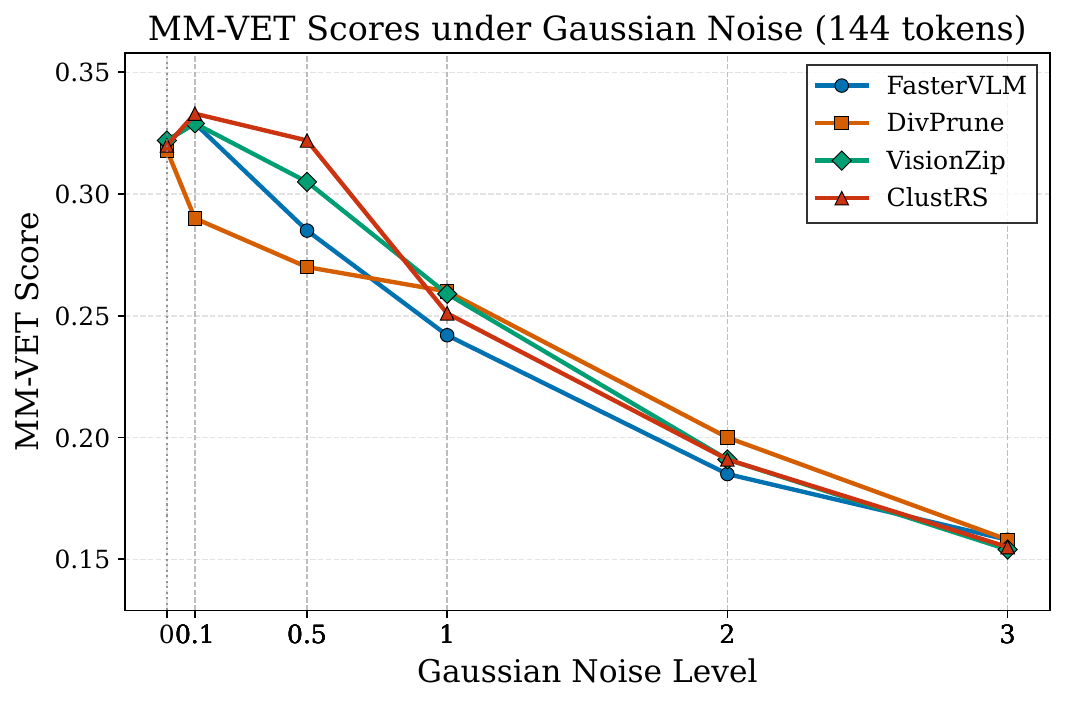}
    \caption*{(c) $k{=}144$ tokens\\[3ex]}
  \end{minipage}
  \hfill
  \begin{minipage}{0.48\linewidth}
    \centering
    \includegraphics[width=\linewidth]{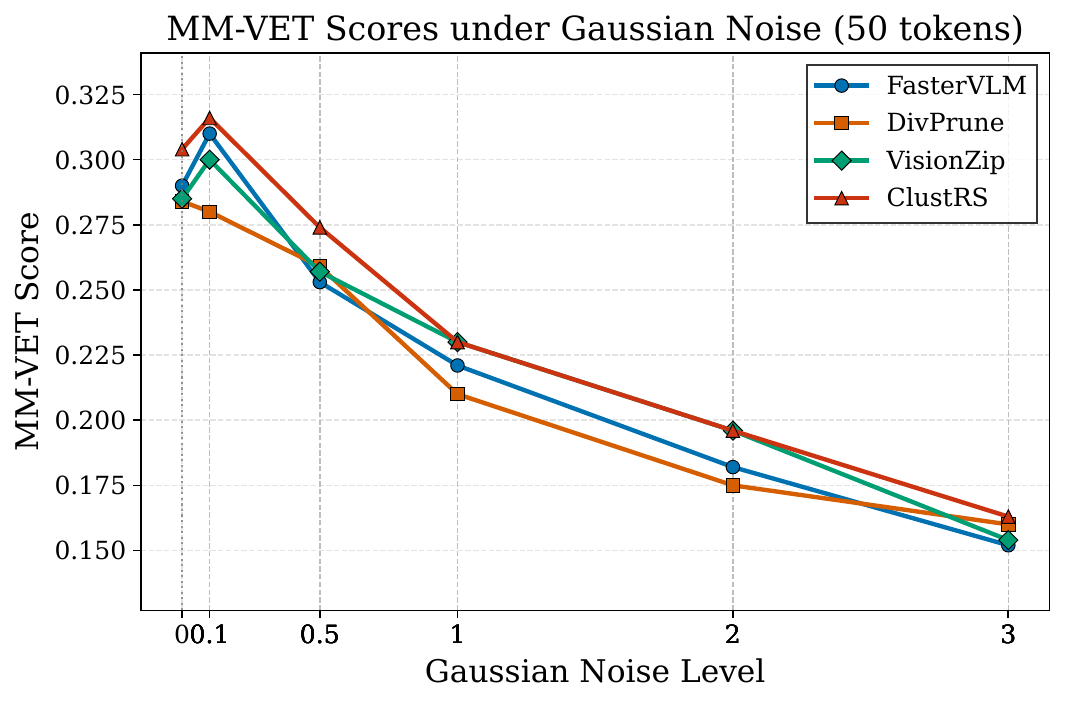}
    \caption*{(b) $k{=}50$ tokens}
  \end{minipage}
  
  \vspace{0.5em}
  
  \begin{minipage}{0.9\linewidth}
    \centering
    \label{fig:severe_gaussian_noise}
    \includegraphics[width=\linewidth]{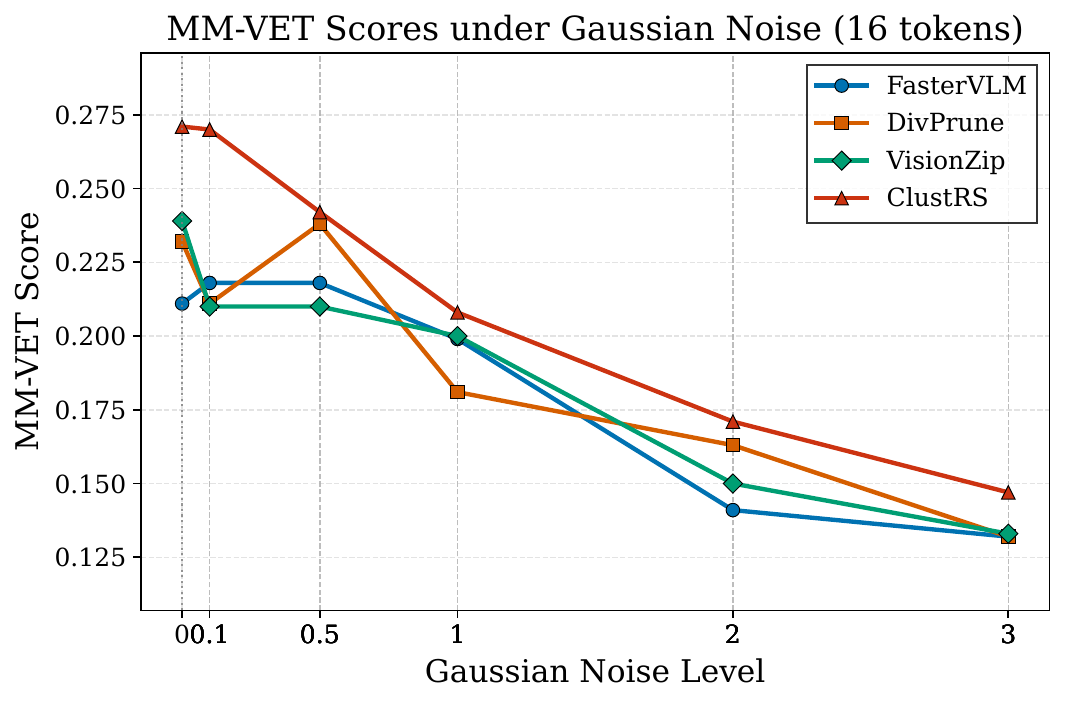}
    \caption*{(a) $k{=}16$ tokens}
  \end{minipage}
  
  \caption{MM-VET accuracy of \textbf{LLaVA-1.5-7B} under additive Gaussian
          noise ($\sigma\!\in\![0.1,3.0]$) for three token budgets
          ($k{=}16,50,144$).\\[1ex]}
  \label{fig:mmvet_gaussian_sweeps}
\end{figure}

\begin{figure}[!htbp]
  \centering
  \begin{minipage}{0.48\linewidth}
    \centering
    \includegraphics[width=\linewidth]{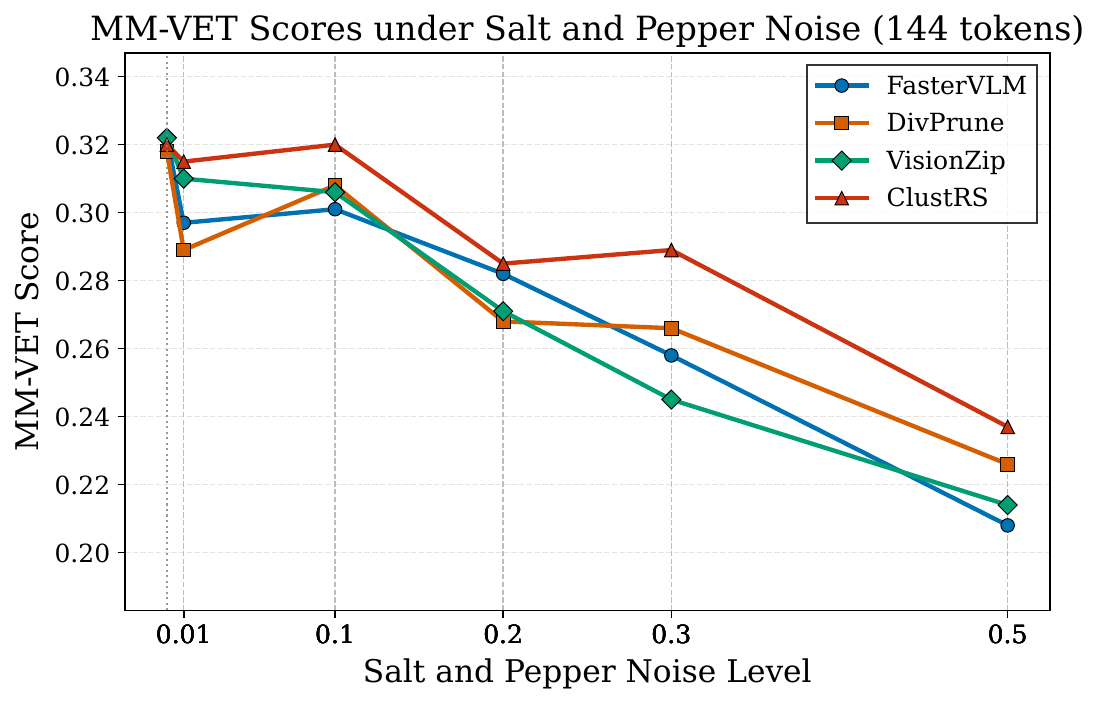}
    \caption*{(c) $k{=}144$ tokens\\[3ex]}
  \end{minipage}
  \hfill
  \begin{minipage}{0.49\linewidth}
    \centering
    \includegraphics[width=\linewidth]{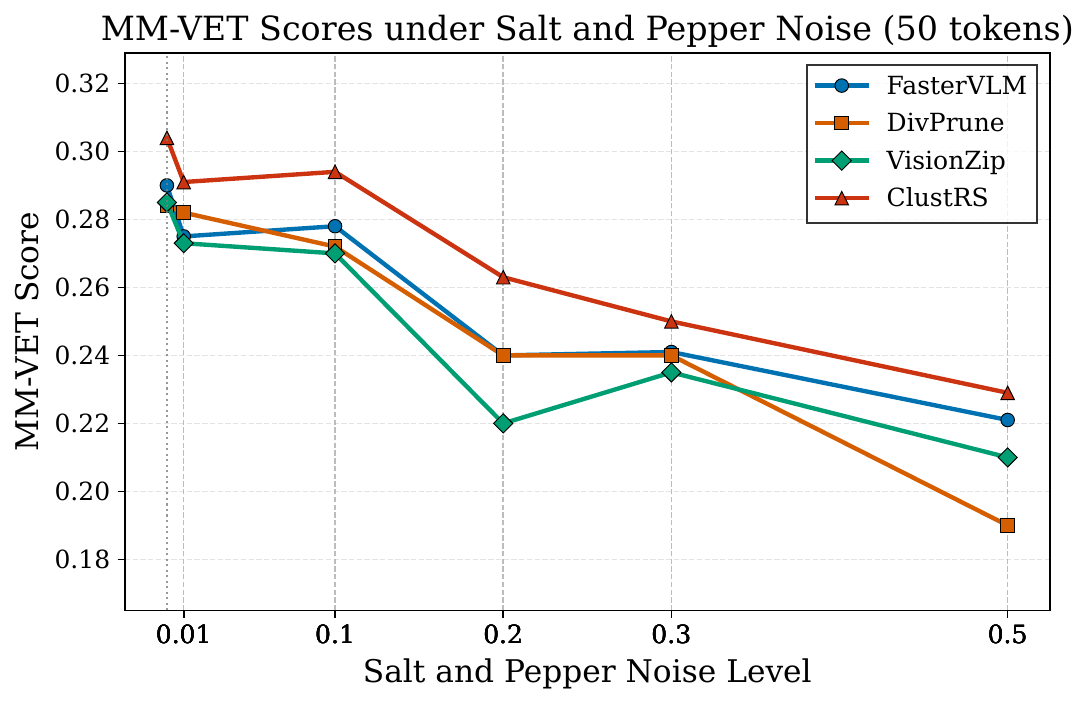}
    \caption*{(b) $k{=}50$ tokens}
  \end{minipage}
  
  \vspace{0.5em}
  
  \begin{minipage}{0.85\linewidth}
    \centering
    
    \includegraphics[width=\linewidth]{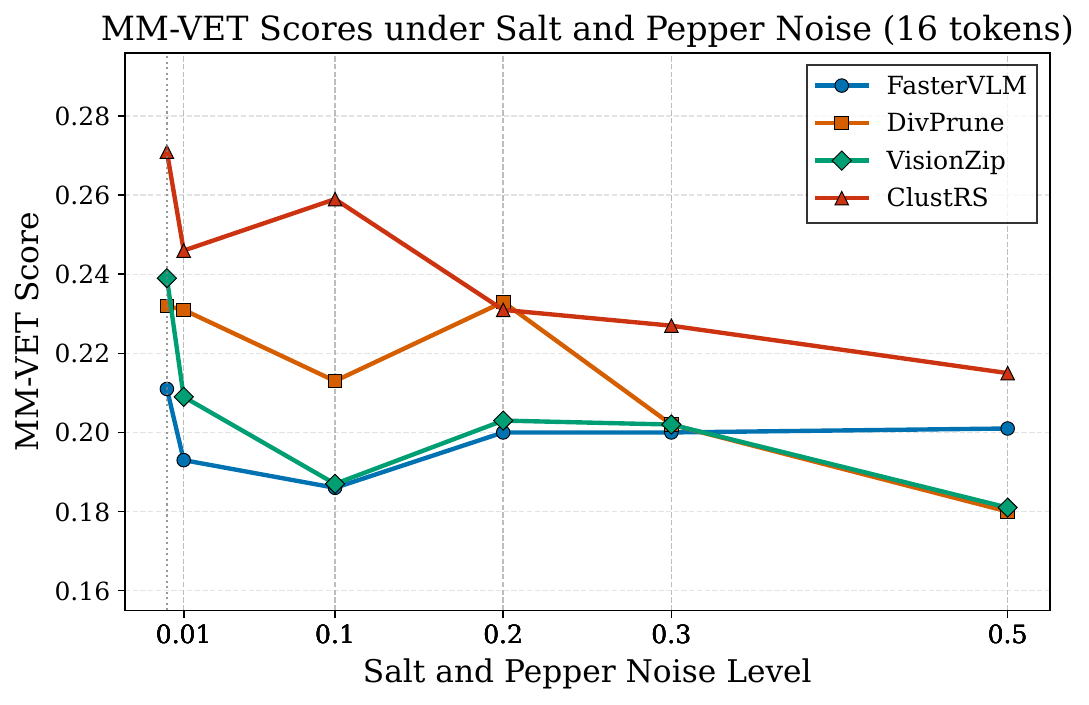}
    \caption*{(a) $k{=}16$ tokens}
  \end{minipage}
  
  \caption{MM-VET accuracy of \textbf{LLaVA-1.5-7B} under \emph{salt–\&–pepper}
          noise ($\rho\!\in\![0.01, 0.50]$) for the three token budgets evaluated in the Chapter. Gaussian-noise counterparts are shown in Figure \ref{fig:mmvet_gaussian_sweeps}.
          }
  \label{fig:mmvet_saltpepper_sweeps} 
\end{figure}

Our evaluation demonstrates that our method ClustRS (C+H selection followed by RRS enhancement) consistently delivers:
\begin{enumerate}
\item Superior performance under clean conditions at low and moderate token budgets;
\item Enhanced robustness to both additive (Gaussian) and impulsive (salt-and-pepper) noise across all token budgets;
\item Particularly significant advantages under extreme compression ($k{=}16$) and high noise conditions.
\end{enumerate}
These improvements come at negligible computational cost: only $\mathcal{O}(kD)$ additional FLOPs per image. This makes our approach ideal for resource-constrained deployment scenarios where both efficiency and robustness are essential. Detailed numerical results are provided in Tables~\ref{tab:mmvet_gaussian_full} and~\ref{tab:mmvet_salt_pepper_full}, and displayed in Figures \ref{fig:mmvet_gaussian_sweeps} and \ref{fig:mmvet_saltpepper_sweeps}.
Table~\ref{tab:additional_corruptions} also shows that this advantage remains on most of the token settings against FasterVLM DivPrune and VisionZip against more naturally occuring types of noise, like Brightness or Cutout.

\begin{table}[!htbp]
\centering
\scriptsize
  
  \begin{tabular}{llcccc}
    \toprule
    Noise $\sigma$ & Token Budget ($k$)
      & FasterVLM & DivPrune & VisionZip & ClustRS \\
    \midrule
    \multirow{3}{*}{Clean}
      & 16  & 0.211 & 0.232 & 0.239 & \textbf{0.271} \\
      & 50  & 0.290 & 0.284 & 0.285 & \textbf{0.304} \\
      & 144 & \textbf{0.322} & 0.318 & \textbf{0.322} & 0.320 \\
    \midrule
    \multirow{3}{*}{0.1}
      & 16  & 0.218 & 0.211 & 0.210 & \textbf{0.270} \\
      & 50  & \textbf{0.310} & 0.280 & 0.300 & \textbf{0.316} \\
      & 144 & 0.329 & 0.290 & 0.329 & \textbf{0.333} \\
    \midrule
    \multirow{3}{*}{0.5}
      & 16  & 0.218 & 0.238 & 0.210 & \textbf{0.242} \\
      & 50  & 0.253 & 0.259 & 0.257 & \textbf{0.274} \\
      & 144 & 0.285 & 0.270 & 0.305 & \textbf{0.322} \\
    \midrule
    \multirow{3}{*}{1.0}
      & 16  & 0.199 & 0.181 & 0.200 & \textbf{0.208} \\
      & 50  & 0.221 & 0.210 & \textbf{0.230} & \textbf{0.232} \\
      & 144 & 0.242 & \textbf{0.260} & 0.259 & 0.251 \\
    \midrule
    \multirow{3}{*}{2.0}
      & 16  & 0.141 & 0.163 & 0.150 & \textbf{0.171} \\
      & 50  & 0.182 & 0.175 & \textbf{0.196} & \textbf{0.196} \\
      & 144 & 0.185 & \textbf{0.200} & 0.191 & 0.191 \\
    \midrule
    \multirow{3}{*}{3.0}
      & 16  & 0.132 & 0.132 & 0.133 & \textbf{0.147} \\
      & 50  & 0.152 & 0.160 & 0.154 & \textbf{0.163} \\
      & 144 & \textbf{0.158} & \textbf{0.158} & 0.154 & 0.155 \\
    \bottomrule
  \end{tabular}
  \caption{MM-VET scores for LLaVA-1.5-7B under Gaussian noise
  with standard deviation $\sigma$. We compare the training-free token
  pruning methods FasterVLM, DivPrune, VisionZip, and ClustRS at different
  retained-token budgets $k$. The best result in each row is shown in bold.}
  \label{tab:mmvet_gaussian_full}

  \bigskip

  \begin{tabular}{llcccc}
    \toprule
    Noise Density ($\rho$) & Token Budget ($k$)
      & FasterVLM & DivPrune & VisionZip & ClustRS \\
    \midrule
    \multirow{3}{*}{0.01}
      & 16  & 0.193 & 0.231 & 0.209 & \textbf{0.246} \\
      & 50  & 0.275 & 0.282 & 0.273 & \textbf{0.291} \\
      & 144 & 0.297 & 0.289 & 0.310 & \textbf{0.315} \\
    \midrule
    \multirow{3}{*}{0.10}
      & 16  & 0.186 & 0.213 & 0.187 & \textbf{0.259} \\
      & 50  & 0.278 & 0.272 & 0.270 & \textbf{0.294} \\
      & 144 & 0.301 & 0.308 & 0.306 & \textbf{0.320} \\
    \midrule
    \multirow{3}{*}{0.20}
      & 16  & 0.200 & \textbf{0.233} & 0.203 & 0.231 \\
      & 50  & 0.240 & 0.240 & 0.220 & \textbf{0.263} \\
      & 144 & 0.282 & 0.268 & 0.271 & \textbf{0.285} \\
    \midrule
    \multirow{3}{*}{0.30}
      & 16  & 0.200 & 0.202 & 0.202 & \textbf{0.227} \\
      & 50  & 0.241 & 0.240 & 0.235 & \textbf{0.250} \\
      & 144 & 0.258 & 0.266 & 0.245 & \textbf{0.289} \\
    \midrule
    \multirow{3}{*}{0.50}
      & 16  & 0.201 & 0.180 & 0.181 & \textbf{0.215} \\
      & 50  & 0.221 & 0.190 & 0.210 & \textbf{0.229} \\
      & 144 & 0.208 & 0.226 & 0.214 & \textbf{0.237} \\
    \bottomrule
  \end{tabular}
  \caption{MM-VET scores for LLaVA-1.5-7B under salt-and-pepper noise
  with corruption density $\rho$. We compare the training-free token
  pruning methods FasterVLM, DivPrune, VisionZip, and ClustRS at different
  retained-token budgets $k$. The best result in each row is shown in bold.}
  \label{tab:mmvet_salt_pepper_full}
\end{table}

\clearpage
\subsection{Performance on ScienceQA-IMG}

\paragraph{Extreme Compression ($k{=}16$)}
Under clean conditions, ClustRS and DivPrune achieve comparable performance (69.01\% and 69.16\% respectively), both outperforming FasterVLM and VisionZip. Our method maintains this advantage under mild noise ($\sigma{=}0.1$), scoring 68.62\% compared to DivPrune's 68.6\% and FasterVLM's 67.48\%. These results on ScienceQA-IMG can be seen in Tables \ref{tab:scienceqa_gaussian_full} and \ref{tab:scienceqa_saltpepper_full}.

\paragraph{Moderate and Conservative Compression ($k\in\{50,144\}$)}
At moderate budgets, ClustRS achieves optimal clean performance (69.21\%, on par with FasterVLM) and maintains superior resilience under mild noise conditions ($\sigma\leq0.5$). With more tokens available ($k{=}144$), our method demonstrates consistent performance across the noise spectrum, achieving 68.77\% under clean conditions and 64.01\% at $\sigma{=}3.0$, outperforming both baselines.

In summary, on ScienceQA-IMG, ClustRS consistently demonstrates competitive or superior performance. While it does not lead in every single experimental condition, it is nearly always the top-performing method and, in the worst case, a close second to the leader. This pattern is understandable given that ScienceQA-IMG is a multiple-choice task less sensitive to the fine-grained visual details where noise resilience becomes critical. Nevertheless, these results complement our MM-VET findings, demonstrating that our approach provides robust and reliable performance across diverse visual reasoning tasks.
\label{sec:scienceqa-results}

\begin{table}[t]
\centering
\small
\begin{tabular}{lcccc}
\toprule
Setting & FasterVLM & DivPrune & VisionZip & ClustRS \\
\midrule
\multicolumn{5}{l}{\textit{$k=16$}} \\
Brightness and contrast & 0.203 & 0.218 & 0.225 & \textbf{0.250} \\
Cutout                 & 0.183 & 0.213 & 0.210 & \textbf{0.214} \\
\midrule
\multicolumn{5}{l}{\textit{$k=50$}} \\
Brightness and contrast & 0.280 & 0.265 & 0.268 & \textbf{0.292} \\
Cutout                 & 0.213 & 0.223 & 0.228 & \textbf{0.233} \\
\midrule
\multicolumn{5}{l}{\textit{$k=144$}} \\
Brightness and contrast & \textbf{0.310} & 0.292 & 0.302 & 0.305 \\
Cutout                 & 0.233 & 0.250 & 0.251 & \textbf{0.267} \\
\bottomrule
\end{tabular}
\caption{Mean MM-VET scores under brightness and contrast transformations and cutout corruption. Brightness-and-contrast results are averaged over six configurations: brightness factors $b\in\{0.5,1.5,2.0\}$ and contrast factors $c\in\{0.5,1.5,2.0\}$. Cutout results are averaged over three severity levels, $p\in\{0.1,0.25,0.4\}$, denoting the fraction of the image area masked.}
\label{tab:additional_corruptions}
\end{table}

\begin{table}[!htbp]
\centering
\scriptsize

  \begin{tabular}{ll cccc}
    \toprule
    Noise ($\rho$) & Token Budget ($k$) & FasterVLM & DivPrune & VisionZip & ClustRS \\
    \midrule
    \multirow{3}{*}{0.0 (Clean)} 
      & 16  & 67.8          & \textbf{69.2} & 68.9 & 69.0           \\
      & 50  & \textbf{69.2} & 68.5          & 68.9 & \textbf{69.2}  \\
      & 144 & 67.9          & 68.3          & 68.4 &  \textbf{68.8}  \\
    \midrule 
    \multirow{3}{*}{0.1}
      & 16  & 67.5          & \textbf{68.6} & 68.3 & \textbf{68.6}  \\
      & 50  & 68.5          & 67.6          & 67.8 & \textbf{68.7}  \\
      & 144 & \textbf{68.4} & 67.6          & 68.0 & 68.0           \\
    \midrule
    \multirow{3}{*}{0.5}
      & 16  & 66.2          & \textbf{67.7} & 67.1 & 67.2           \\
      & 50  & 67.8          & 67.7          & 67.6 & \textbf{68.1}  \\
      & 144 & 67.4          & \textbf{68.1} & 67.5 & 67.7           \\
    \midrule
    \multirow{3}{*}{1.0}
      & 16  & 63.9          & \textbf{64.9} & 64.5 & \textbf{64.9}  \\
      & 50  & 64.3          & \textbf{64.8} & 64.6 & 64.2           \\
      & 144 & 64.1          & 64.3          & 64.2 & \textbf{64.4}  \\
    \midrule
    \multirow{3}{*}{2.0}
      & 16  & \textbf{64.0} & 63.9          & 63.1 & 63.6           \\
      & 50  & 63.5          & \textbf{64.0} & 63.5 & 63.8           \\
      & 144 & \textbf{63.8} & 63.6          & 63.5 & \textbf{63.8}  \\
    \midrule
    \multirow{3}{*}{3.0}
      & 16  & 63.6          & 63.7          & 63.6 & \textbf{63.8}  \\
      & 50  & 63.2          & \textbf{64.1} & 63.6 & 63.5           \\
      & 144 & 63.8          & 63.6          & 63.8 & \textbf{64.0}  \\
    \bottomrule
  \end{tabular}
  \caption{ScienceQA scores under Gaussian noise for three token budgets $k\in\{16,50,144\}$.}
  \label{tab:scienceqa_gaussian_full}
  
  \bigskip

  \begin{tabular}{ll cccc}
    \toprule
    Noise ($\rho$) & Token Budget ($k$) & FasterVLM & DivPrune & VisionZip & ClustRS \\
    \midrule
    \multirow{3}{*}{0.01} 
      & 16  & 67.5          & 67.7          & 67.6 & \textbf{68.6} \\
      & 50  & \textbf{68.3} & \textbf{68.3} & 67.8 & 67.9          \\
      & 144 & 67.4          & 67.5          & 67.5 & \textbf{67.6} \\
    \midrule
    \multirow{3}{*}{0.10}
      & 16  & 68.4          & \textbf{68.6} & 68.0 & 68.1          \\
      & 50  & 68.5          & 67.1          & 68.3 & \textbf{68.6} \\
      & 144 & \textbf{68.4} & 67.8          & 68.0 & 68.0          \\
    \midrule
    \multirow{3}{*}{0.20}
      & 16  & 68.7          & 68.2          & 68.4 & \textbf{68.8} \\
      & 50  & \textbf{68.8} & 67.8          & 68.1 & 68.1          \\
      & 144 & \textbf{68.7} & 68.0          & 68.1 & 67.7          \\
    \midrule
    \multirow{3}{*}{0.30}
      & 16  & 67.0          & 67.5          & 67.3 & \textbf{67.6} \\
      & 50  & 68.7          & 68.2          & 68.4 & \textbf{69.3} \\
      & 144 & \textbf{68.6} & 68.3          & 67.4 & 68.5          \\
    \midrule
    \multirow{3}{*}{0.50}
      & 16  & 64.9          & 66.0          & 65.1 & \textbf{66.7} \\
      & 50  & \textbf{65.9} & \textbf{65.9} & 65.2 & 65.6          \\
      & 144 & \textbf{65.7} & 65.6          & 65.4 & 65.4          \\
    \bottomrule
  \end{tabular}
  \caption{ScienceQA scores under Salt\&Pepper noise for three token budgets $k\in\{16,50,144\}$.}
  \label{tab:scienceqa_saltpepper_full}
\end{table}

\subsection{The impact of our algorithm on runtime performance}
The overhead of the proposed method was benchmarked on LLaVA-1.5-7b and, more importantly, contextualized it within the end-to-end inference time. First, we measured the 'visual overhead' (Vision Encoder + Token Pruning Method) on an A100 GPU with 40Gb of memory, running at a base frequency of 1.4GHz, capable of delivering 312 TFLOPS. While this high-end hardware does not represent a constrained device, it nonetheless is useful to benchmark the relative computational cost of the pruning algorithms themselves. Our more sophisticated robust clustering adds a negligible, one-time cost compared to the baselines, that can be seen in Table \ref{tab:visual_overhead_comparison}.

As shown, our method at k=16 (the configuration we focus on) adds only 1ms of overhead compared to the fastest baseline. Although it is true, our algorithm is the only one to increase in complexity with the number of tokens, resulting in a 50\% increase in runtime at 144 tokens. However, we argue this is negligible for two reasons: 
\begin{enumerate}
    \item The LLM is usually larger than the vision part.
    \item The vision part is only needed once, whereas the (autoregressive) LLM part is called for every output token generated.    
\end{enumerate}
To quantify this, we measured the end-to-end generation time on MM-VET:
\begin{itemize}
    \item Generating a short 14-token answer takes 355ms.
    \item Generating a medium, 64-token answer takes ~2000ms (2.0 seconds).
\end{itemize}
So this one-time pruning overhead of 35ms represents only ~1.6\% of the total time for a 64-token answer, which is negligible as expected.

\begin{table}
\centering
\begin{tabular}{|l|c|c|}
\hline
\textbf{Method} & \textbf{k (Tokens Kept)} & \textbf{Total Visual Overhead (ms)} \\
\hline
FasterVLM       & 144 and 16               & 33-34                                  \\
DivPrune        & 144 and 16               & 39-40    
    \\
VisionZip        & 144 and 16               & 34-38                                \\
ClustRS (Ours)  & 16                       & 35                                  \\
ClustRS (Ours)  & 144                      & 52                                  \\
\hline
\end{tabular}
\caption{Comparison of total visual overhead for different token pruning methods.}
\label{tab:visual_overhead_comparison}
\end{table}

\begin{table}
\centering
\scriptsize
\begin{tabular}{|l|l|c|c|c|}
\hline
\textbf{Corruption Type} & \textbf{Level} & \textbf{FasterVLM} & \textbf{DivPrune} & \textbf{ClustRS (Ours)} \\
\hline
Motion Blur    & High (B=31)   & 0.184 & 0.219 & \textbf{0.222} \\
Motion Blur    & Medium (B=15) & 0.218 & 0.235 & \textbf{0.254} \\
JPEG Artifacts & Medium (q=60) & 0.250 & 0.261 & \textbf{0.284}\\
JPEG Artifacts & Low (q=30)    & 0.240 & 0.230 & \textbf{0.282} \\
\hline
\end{tabular}
\caption{MM-VET scores of LLaVA-1.5-7b with 16 visual tokens under different image corruptions. Here B is the size of the blurring kernel, and q is the quality factor of the PIL library (going from 1 -worst- to 95 -best-)}
\label{tab:corruption_performance}
\end{table}

\section{Analysis and discussion}

\subsection{Token selection and enhancement: a complementary approach}

Our approach combines robust token selection (C+H) with subsequent token enhancement (RRS). C+H provides a noise-aware selection by balancing attention-guided salience with clustering-enforced diversity, stabilizing choices against noise. The complementary nature of adding RRS is quantified in our ablation study (Table\ref{tab:ablation_c_h_rrs_noisy}).
The results show that while C+H establishes a strong baseline, RRS offers further significant gains, especially when the token budget is severely constrained (k=16). In this regime, RRS consistently improves accuracy under all tested noise levels (e.g., +0.037 at $\sigma=3.0$), demonstrating its value in refining the few, precious tokens.

\begin{table}
  \centering

  \setlength{\tabcolsep}{5pt}
  \begin{tabular}{ll ccc c}
    \toprule
    Tokens & Noise            & Clustering        & C+H            & ClustRS & Gain  \\
    ($k$)  & ($\sigma$)       &               & only           &                     & (RRS) \\
    \midrule
    \multirow{4}{*}{16} 
        & 0.5              & 0.233          & 0.254          & \textbf{0.272}      & +0.018 \\
        & 1.0              & 0.195          & 0.223          & \textbf{0.250}      & +0.027 \\
        & 2.0              & 0.185          & 0.193          & \textbf{0.222}      & +0.029 \\
        & 3.0              & 0.151          & 0.156          & \textbf{0.193}      & +0.037 \\
    \midrule
    \multirow{4}{*}{50} 
        & 0.5              & 0.284          & \textbf{0.312} & 0.290               & -0.022 \\ 
        & 1.0              & 0.244          & 0.270          & \textbf{0.273}      & +0.003 \\
        & 2.0              & 0.188          & 0.207 & \textbf{0.215}      & +0.008 \\ 
        & 3.0              & 0.176          & \textbf{0.198} & 0.169               & -0.029 \\ 
    \midrule
    \multirow{4}{*}{144} 
        & 0.5              & 0.310          & 0.293 & \textbf{0.312}      & +0.019 \\ 
        & 1.0              & 0.265          & 0.271          & \textbf{0.271}      & ~0.000 \\
        & 2.0              & 0.192          & 0.198          & \textbf{0.203}      & +0.005 \\
        & 3.0              & 0.170          & \textbf{0.186} & 0.177               & -0.009 \\ 
    \bottomrule
  \end{tabular}
  
  \caption{Ablation study on MM-VET under Gaussian noise, showing the distinct contributions of our proposed components. We compare standard k-means++ Clustering (\textbf{Clustering}), robust clustering with Huber distance (\textbf{C+H}), and our full \textbf{ClustRS} method which adds Residual Shrinkage (RRS). The results demonstrate that while clustering provides a baseline improvement, the Huber distance is critical for robustness, and RRS provides further gains, especially in low-token regimes. All scores are accuracy on the LLaVA-1.5-7B model.}
  \label{tab:ablation_c_h_rrs_noisy}
\end{table}

\paragraph{Diminishing Returns with Larger Budgets}
However, the benefits of both aggressive clustering (C+H) and residual shrinkage (RRS) can diminish as the token budget increases. The aggressive nature of selecting only one representative per cluster in C+H might discard tokens carrying fine-grained, complementary details from nearby patches that a less restrictive selector might keep. Similarly, RRS provides less pronounced benefits when more tokens are available, and in some cases, C+H alone can perform slightly better (Table~\ref{tab:ablation_c_h_rrs_noisy} for k=50,144 at $\sigma=3.0$). This suggests that while our ClustRS pipeline excels in high-compression, noisy scenarios critical for embedded systems, its aggressive optimization for these conditions may be less crucial when constraints are relaxed.

\subsection{Is the spatial information useful ?}
Paper \cite{xing2025largevisionlanguagemodelslook} by Xing et al. serves as an inspiration for our work's approach to token pruning. Their analysis of where VLMs look when answering questions reveals key insights about the behaviors of these models. Their findings show that important visual information is often distributed inconsistently across spatial locations, with models sometimes producing correct answers while focusing on seemingly irrelevant image regions. This suggests that the spatial arrangement of visual tokens may not be as important as their semantic content. Based on this observation, our approach deliberately moves away from spatial-based token pruning strategies that preserve tokens based on their location in the image grid. Instead, we adopt a more semantic and information-centric approach that evaluates tokens based on their contribution to the model's overall understanding, regardless of their spatial position. By focusing on the information content rather than spatial arrangement, our method can better preserve the tokens most critical for the model's reasoning process, even when these tokens are scattered across different regions of the image.

\section{Conclusion}
\label{sec:conclusion}

We introduced ClustRS, a method made of two complementary, training-free techniques for efficient and robust token selection in VLMs. C+H effectively selects diverse and salient tokens, while RRS enhances robustness by adaptively reducing noise within these tokens.

Our experiments demonstrate that the combination of C+H and RRS consistently outperforms existing current training-free pruning methods on benchmarks such as MM-VET and ScienceQA-IMG, particularly under severe noise conditions and aggressive token pruning scenarios. Additionally, our method successfully adapts to newer SigLIP-based backbones, highlighting its versatility, though it needed an analysis of the differences between CLIP and SigLIP. Looking ahead, while token selection methods like C+H achieve robust and efficient representations, there remains significant potential to refine denoising and token synthesis further. Future research should explore advanced token refinement strategies, including lightweight learning-based denoising or generative methods, to enhance VLM robustness and deployment efficiency even more. A remaining limitation is that feature-space clustering may discard spatially distinct but semantically similar tokens, which could affect fine-grained localization and spatial-relation reasoning; evaluating and mitigating this behavior is an important direction for future work.



\clearpage
\appendix
\section{Adaptation to \textit{LLaVA-OneVision}}
\addcontentsline{toc}{section}{Appendix A: OneVision Results}

\noindent
This appendix documents how the training-free token-selection pipeline
of the main paper transfers from the CLIP-based \textit{LLaVA-1.5} to
the newer SigLIP-based \textit{LLaVA-OneVision}.  We first introduce the norm-aware salience modification that restores a rich ranking
signal for SigLIP tokens, then report full MM-VET results.

\subsection{Adapting C\,+\,H to \textit{LLaVA-OneVision}: Norm-Aware Salience}
\label{sec:salience}

\vspace{2pt}\noindent
\textbf{Why a new salience signal?}  
The attention maps produced by the SigLIP vision backbone in \textit{LLaVA-OneVision} differ markedly from those of the CLIP backbone used in \textit{LLaVA-1.5}.  In CLIP, the self-attention weights of the \texttt{[CLS]} token form a highly peaked distribution: the strongest token typically receives a score about $5\times$ larger than the median, giving the clustering stage a clear, high-contrast ranking signal.  SigLIP, by contrast, yields a much flatter distribution in which the top weight is only $\sim\!2\times$ the median.  With such weak contrast, our original C\,+\,H selector struggles: attentional weights become almost uniform, so the $k$-means++ seeding and subsequent representative pick tend to favour whatever tokens happen to be most numerous, often low-information background patches.

At the same time, SigLIP encodes visual importance in a different channel: the length of each feature vector.  Salient foreground patches exhibit noticeably larger norms, spanning $\|v_i\|\!\in[1.8,2.1]$, whereas uninformative background tokens cluster around $\|v_i\|\!\approx1.0$.  (CLIP embeddings, in comparison, are nearly length-normalised to $[0.98,1.05]$.)  By incorporating this norm as a small “bonus” term in the salience score we effectively restore the missing dynamic range: tokens that are both attended to and long receive markedly higher scores than the background.  This hybrid signal re-establishes a decisive ordering, allowing C\,+\,H to place centroids on truly informative regions and to choose meaningful cluster representatives, recovering the robustness and diversity we observe on CLIP-based models.

\paragraph{Norm-aware salience score}
We inject this additional cue through a lightweight
norm bonus added to the raw attention:
\begin{equation}
\label{eq:salience}
s_i = a_i + \lambda \max(0, |v_i| - \mu_{|v|})
\end{equation}
where $a_i$ is the SigLIP [CLS] attention,
$\mu_{\|v\|}$ is the mean token norm of the image,
and $\lambda=2\,\bar a$ with $\bar a=\tfrac1N\sum_i a_i$.
Equation~\eqref{eq:salience} preserves the original scale
($\mathbb{E}[s_i]\!\approx\!\mathbb{E}[a_i]$) while amplifying
tokens whose norms lie above average.

\paragraph{Drop-in replacement inside C\,+\,H :}
We simply replace every occurrence of $a_i$ in
Section~4.1 by $s_i$:
\begin{enumerate}[leftmargin=1.5em,itemsep=2pt]
  \item[$\triangleright$] K-means++ seeding:
        $p_i \propto s_i\,d(\mathbf v_i,\mathbf C)^2$.
  \item[$\triangleright$] Centroid updates:
        weights $w_i = s_i / \sum_{m:c(m)=j} s_m$.
  \item[$\triangleright$] Cluster representative:
        $\operatorname*{arg\,max}_{i\in\mathcal{C}_j} s_i$.
\end{enumerate}

This one-line modification restores a dynamic range of ${\sim}4{\times}$ between top- and median-ranked tokens, allowing C\,+\,H to match the MM-VET accuracy of FasterVLM and DivPrune on \textit{OneVision} while using \textbf{only 50 tokens versus their 180} under mild noise ($\sigma{=}0.5$).

The norm-aware term acts only as an additional salience cue and does not directly determine token retention. Feature-space clustering continues to enforce representational coverage, allowing low-norm background tokens to be retained when they encode contextual information that is distinct from the selected foreground tokens.

\paragraph{Compatibility}
The norm-aware salience term is:
(i) \textbf{training-free};
(ii) adds \(\mathcal{O}(N)\) FLOPs (one norm per token);
(iii) leaves the Huber distance, clustering loop, and optional
RRS step untouched.

\subsection{MM-VET Results on OneVision}
\label{app:onevision-mmvet}


\begin{table*}[!ht]
  \centering
  
  \label{tab:mmvet_gaussian_condensed}
  \scriptsize 
  \begin{tabular}{ll ccc}
    \toprule
    Noise ($\sigma$) & Token Budget ($k$) & FasterVLM & DivPrune & ClustRS (salience) \\
    \midrule
    \multirow{3}{*}{0.0 (Clean)} 
      & 16  & 0.138          & 0.164 & \textbf{0.272} \\
      & 50  & 0.145          & 0.257 & \textbf{0.347} \\
      & 180 & 0.293          & 0.334 & \textbf{0.418} \\
    \midrule 
    \multirow{3}{*}{0.1}
      & 16  & 0.123          & 0.166 & \textbf{0.284} \\
      & 50  & 0.130          & 0.215 & \textbf{0.348} \\
      & 180 & 0.296          & 0.289 & \textbf{0.363} \\
    \midrule
    \multirow{3}{*}{0.5}
      & 16  & 0.133          & 0.146 & \textbf{0.215} \\
      & 50  & 0.131          & 0.175 & \textbf{0.254} \\
      & 180 & 0.193          & 0.252 & \textbf{0.277} \\
    \midrule
    \multirow{3}{*}{1.0}
      & 16  & 0.131          & 0.136 & \textbf{0.153} \\
      & 50  & 0.114          & 0.163 & \textbf{0.171} \\
      & 180 & 0.156          & 0.155 & \textbf{0.198} \\
    \midrule
    \multirow{3}{*}{2.0}
      & 16  & 0.119          & 0.115 & \textbf{0.150} \\
      & 50  & \textbf{0.128} & 0.120 & 0.115          \\
      & 180 & 0.128          & 0.138 & \textbf{0.147} \\
    \midrule
    \multirow{3}{*}{3.0}
      & 16  & 0.119          & 0.110 & \textbf{0.133} \\
      & 50  & \textbf{0.122} & 0.117 & 0.113          \\
      & 180 & 0.117          & 0.121 & \textbf{0.126} \\
    \bottomrule
  \end{tabular}
  \caption{MM-VET scores under Gaussian noise ($\sigma$) for LLaVA OneVision 7b across different token budgets ($k$). The $\sigma=0.0$ row represents performance on clean images. Scores compare FasterVLM, DivPrune, and our combined C+H (salience) + RRS method.}

  \bigskip

  \label{tab:mmvet_saltpepper_full_onevision}
  \begin{tabular}{ll ccc}
    \toprule
    Noise ($\rho$) & Token Budget ($k$) & FasterVLM & DivPrune & ClustRS (salience) \\
    \midrule
    \multirow{3}{*}{0.01}
      & 16  & 0.124 & 0.177 & \textbf{0.282} \\
      & 50  & 0.139 & 0.243 & \textbf{0.334} \\
      & 180 & 0.305 & 0.327 & \textbf{0.391} \\
    \midrule
    \multirow{3}{*}{0.10}
      & 16  & 0.142 & 0.153 & \textbf{0.254} \\
      & 50  & 0.146 & 0.217 & \textbf{0.354} \\
      & 180 & 0.254 & 0.297 & \textbf{0.343} \\
    \midrule
    \multirow{3}{*}{0.20}
      & 16  & 0.121 & 0.166 & \textbf{0.232} \\
      & 50  & 0.131 & 0.229 & \textbf{0.279} \\
      & 180 & 0.259 & 0.265 & \textbf{0.340} \\
    \midrule
    \multirow{3}{*}{0.30}
      & 16  & 0.129 & 0.160 & \textbf{0.208} \\
      & 50  & 0.132 & 0.185 & \textbf{0.239} \\
      & 180 & 0.241 & 0.254 & \textbf{0.326} \\
    \midrule
    \multirow{3}{*}{0.50}
      & 16  & 0.130 & 0.137 & \textbf{0.163} \\
      & 50  & 0.125 & 0.186 & \textbf{0.206} \\
      & 180 & 0.178 & 0.229 & \textbf{0.244} \\
    \bottomrule
  \end{tabular}
  \caption{MM-VET scores under Salt\&Pepper noise for three token budgets $k\in\{16,50,180\}$.}
\end{table*}

\begin{figure*}
  \centering
  
  \begin{minipage}{0.48\linewidth}
    \centering
    \includegraphics[width=\linewidth]{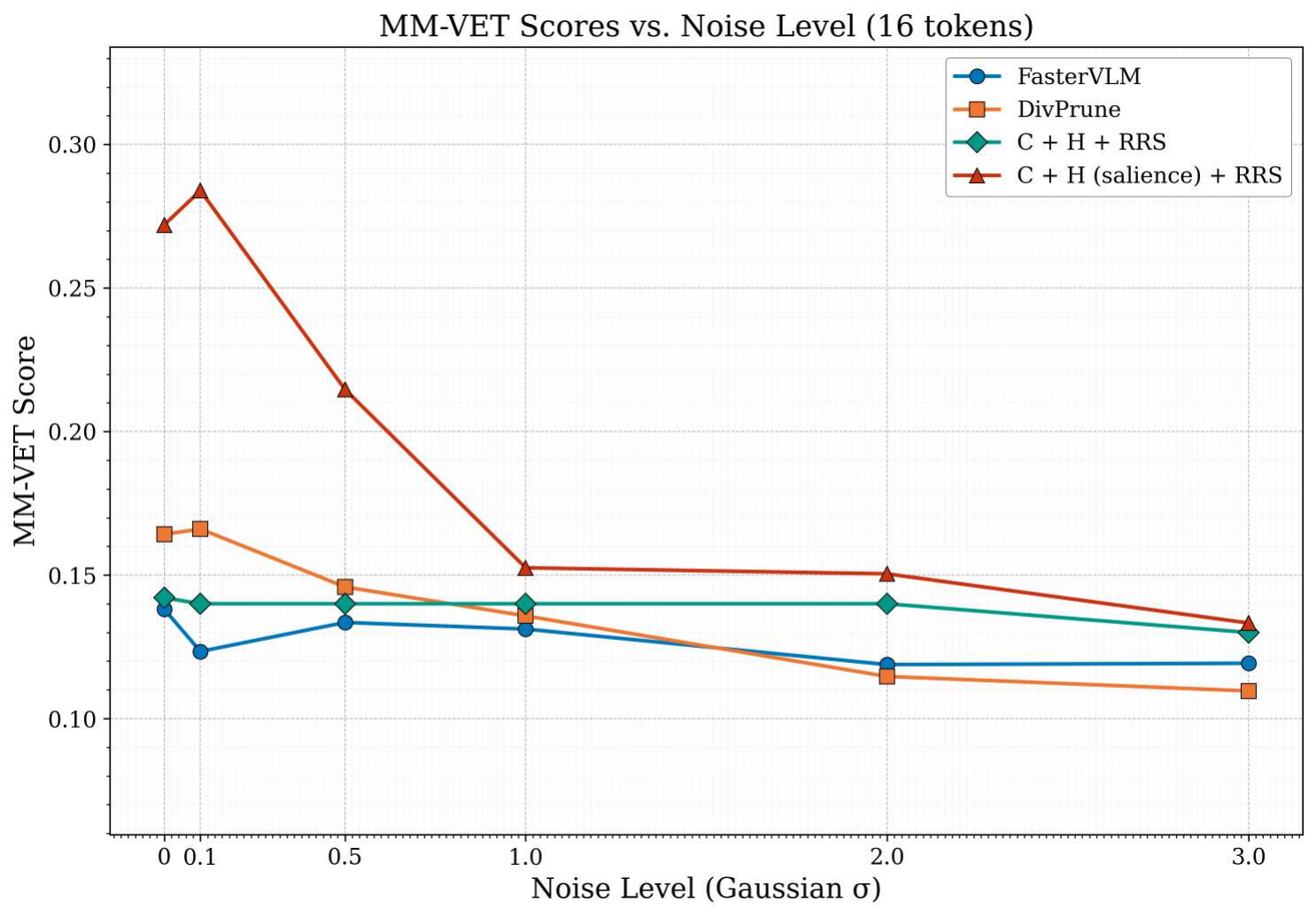}
    \caption*{(a) $k{=}16$ tokens}
  \end{minipage}
  \hfill
  \begin{minipage}{0.48\linewidth}
    \centering
    \includegraphics[width=\linewidth]{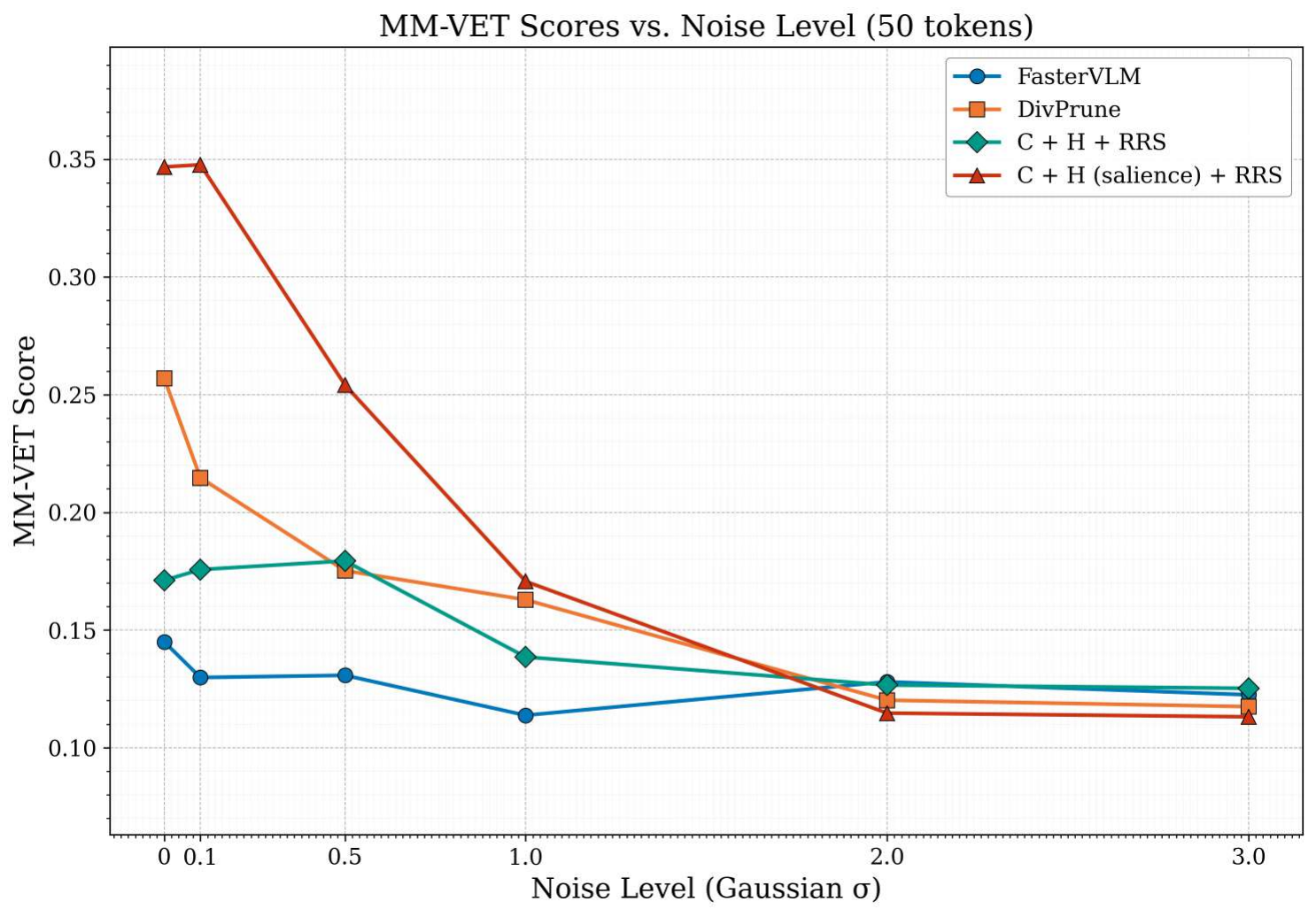}
    \caption*{(b) $k{=}50$ tokens}
  \end{minipage}
  
  \vspace{0.5em}

  \begin{minipage}{0.48\linewidth}
    \centering
    \includegraphics[width=\linewidth]{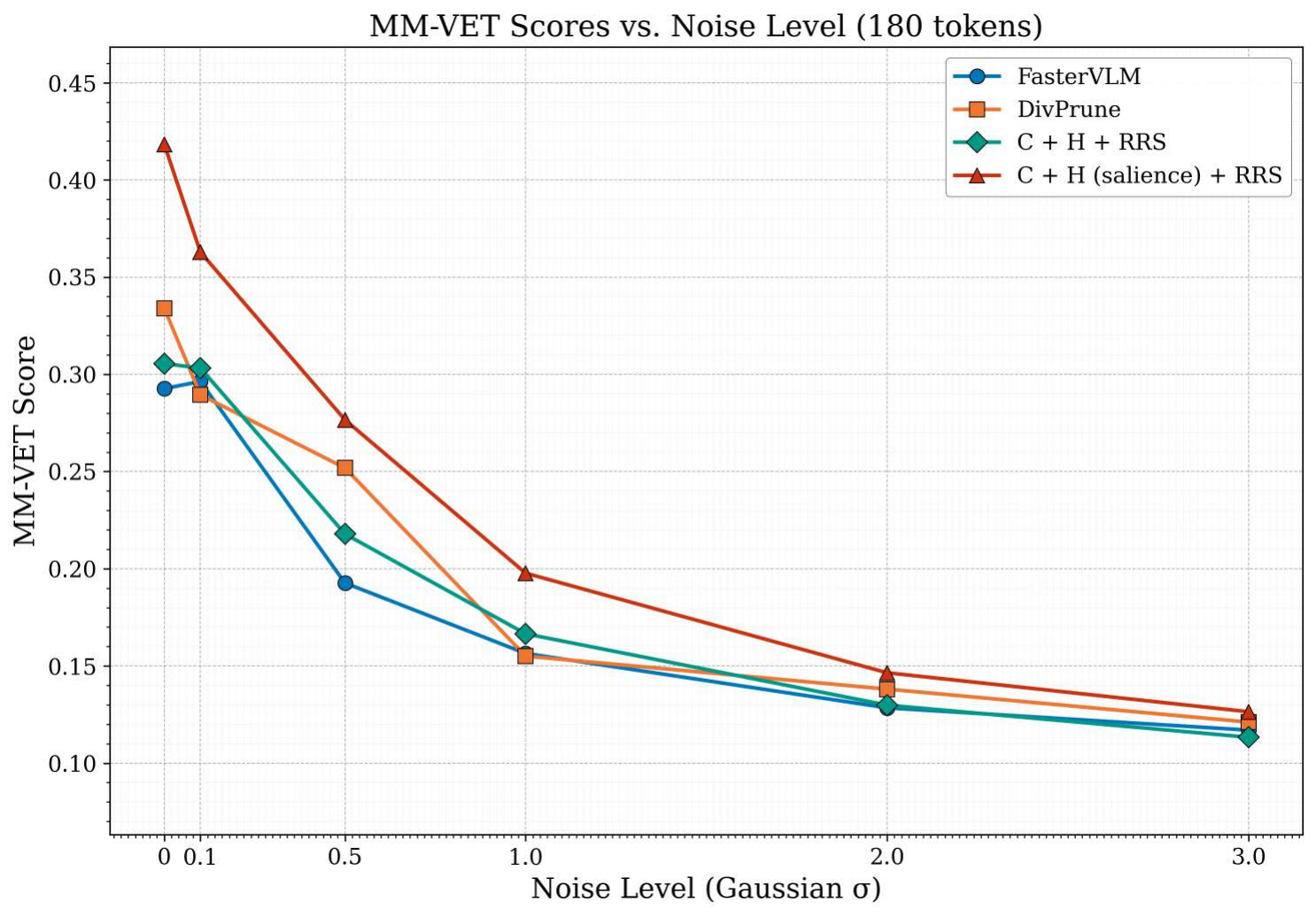}
    \caption*{(c) $k{=}180$ tokens\\[3ex]}
  \end{minipage}
  
  \caption{MM-VET accuracy of \textbf{LLaVA 7b Onevision} under additive Gaussian noise ($\sigma\!\in\![0.1,3.0]$) for token budgets $k\in\{16,50,180\}$.\\[1ex]}
  \label{fig:mmvet_gaussian_sweeps_onevision}
\end{figure*}

\paragraph{Extreme Compression Regime (k=16)}
At the k=16 budget, the impact of the norm-aware salience term is stark. The adapted ClustRS method achieves a clean accuracy of 0.272, dramatically outperforming FasterVLM (0.138, +13.4 pp), DivPrune (0.164, +10.8 pp), and the original ClustRS without salience (0.142, +13.0 pp). This highlights the necessity of the adaptation for SigLIP. Under heavy Gaussian noise ($\sigma=3.0$), the salience-aware method maintains the highest accuracy (0.133), leading FasterVLM (0.119) and DivPrune (0.110). This robustness extends to Salt \& Pepper noise; at the highest tested level ($\rho=0.5$), ClustRS scores 0.163, clearly surpassing FasterVLM (0.130, +3.3 pp) and DivPrune (0.137, +2.6 pp).
\paragraph{High Compression Regime (k=50)}
With k=50 tokens, the ClustRS method continues to dominate in clean conditions, reaching 0.347 accuracy – a massive improvement over FasterVLM (0.145, +20.2 pp) and DivPrune (0.257, +9.0 pp). While performance under extreme Gaussian noise ($\sigma=3.0$) dips slightly (0.113), the method demonstrates strong resilience to Salt \& Pepper noise. At $\rho=0.5$, ClustRS achieves 0.206 accuracy, substantially outperforming both FasterVLM (0.125, +8.1 pp) and DivPrune (0.186, +2.0 pp), confirming its robustness advantage against impulsive noise at this budget.
\paragraph{Low Compression Regime (k=180)}
Increasing the budget to k=180 (the same amount of token reduction, 75\%, for SigLIP as 144 was for CLIP), the ClustRS approach yields the best clean performance at 0.418 accuracy, a considerable gain over FasterVLM (0.293, +12.5 pp) and DivPrune (0.334, +8.4 pp). This method also shows superior robustness at this budget. Under heavy Gaussian noise ($\sigma=3.0$), it retains the top performance (0.126) compared to FasterVLM (0.117) and DivPrune (0.121). Similarly, under high Salt \& Pepper noise ($\rho=0.5$), ClustRS scores 0.244, significantly higher than FasterVLM (0.178, +6.6 pp) and DivPrune (0.229, +1.5 pp). This confirms that with a sufficient token budget, the salience-aware selection provides robust performance across different noise types for OneVision.

\newpage
\section{Sensitivity analysis}
\label{sec:sensitivity}

In this appendix, we analyze the sensitivity of ClustRS to the Huber threshold $\tau$ and to the clipping interval applied to the RRS shrinkage coefficient $\alpha_i$. All other experimental settings are kept identical to those used in the main experiments. The default values, $\tau=1$ and $\alpha_i\in[0.10,0.95]$, are shared across all token budgets and corruption conditions and are not tuned separately for each setting.

Table~\ref{tab:tau_sensitivity} reports the effect of varying the Huber threshold. Performance remains relatively stable for $\tau\in[0.5,2]$, with mean scores ranging from $0.257$ to $0.261$. The default value $\tau=1$ achieves the highest overall mean and performs best or ties for the best result in most clean, mildly corrupted, and salt-and-pepper settings. Under severe Gaussian corruption, however, the optimal value depends on the token budget, indicating that stronger robustness is not uniformly obtained from a single threshold.

Table~\ref{tab:alpha_sensitivity} evaluates the clipping interval applied to $\alpha_i$. When RRS is used without effective clipping, corresponding to $[0,1]$, performance decreases substantially. In contrast, the moderate clipping intervals $[0.05,0.95]$, $[0.10,0.95]$, and $[0.20,0.90]$ produce similar mean scores, indicating that performance is relatively insensitive to the precise clipping bounds within this range. The default interval $[0.10,0.95]$ is therefore retained as a shared setting.

The comparison with C+H without RRS also shows that the benefit of residual shrinkage depends on the operating condition. RRS generally improves performance under mild Gaussian and salt-and-pepper corruption, but it can reduce performance in some severe Gaussian-noise settings. This suggests that the reliability of the cluster centroid and the appropriate shrinkage strength vary with both the corruption level and the token budget. An adaptive mechanism for activating or calibrating RRS is an interesting direction for future work.

\begin{table*}[t]
  \centering
  \scriptsize
  \setlength{\tabcolsep}{7pt}
  \begin{tabular}{llccccc}
    \toprule
    Condition & Token Budget ($k$)
      & $\tau=0.25$ & $\tau=0.5$ & $\tau=1.0$
      & $\tau=2.0$ & $\tau=4.0$ \\
    \midrule
    \multirow{4}{*}{$k=16$}
      & Clean
      & 0.250 & 0.260 & \textbf{0.273} & 0.261 & 0.259 \\
      & Gaussian ($\sigma=0.5$)
      & 0.220 & \textbf{0.240} & \textbf{0.240} & 0.230 & 0.220 \\
      & Gaussian ($\sigma=2.0$)
      & 0.167 & \textbf{0.173} & 0.161 & 0.165 & 0.157 \\
      & Salt-and-pepper ($\rho=0.10$)
      & 0.230 & 0.252 & \textbf{0.256} & 0.248 & 0.238 \\
    \midrule
    \multirow{4}{*}{$k=50$}
      & Clean
      & 0.283 & 0.294 & 0.302 & \textbf{0.307} & 0.298 \\
      & Gaussian ($\sigma=0.5$)
      & 0.252 & 0.260 & \textbf{0.270} & 0.266 & 0.254 \\
      & Gaussian ($\sigma=2.0$)
      & 0.181 & \textbf{0.207} & 0.190 & 0.192 & 0.181 \\
      & Salt-and-pepper ($\rho=0.10$)
      & 0.273 & 0.284 & \textbf{0.297} & 0.292 & 0.278 \\
    \midrule
    \multirow{4}{*}{$k=144$}
      & Clean
      & 0.300 & \textbf{0.310} & \textbf{0.310} & 0.309 & 0.300 \\
      & Gaussian ($\sigma=0.5$)
      & 0.298 & 0.313 & \textbf{0.321} & 0.318 & 0.295 \\
      & Gaussian ($\sigma=2.0$)
      & 0.189 & 0.190 & \textbf{0.191} & 0.180 & 0.171 \\
      & Salt-and-pepper ($\rho=0.10$)
      & 0.308 & 0.324 & \textbf{0.325} & 0.311 & 0.304 \\
    \midrule
    \multicolumn{2}{l}{Mean}
& 0.246 & 0.259 & \textbf{0.261} & 0.257 & 0.246 \\
    \bottomrule
  \end{tabular}
  \caption{Sensitivity analysis of the Huber threshold $\tau$ on MM-VET with LLaVA-1.5-7B. The default value is $\tau=1$. We see a broad performance plateau for $\tau\in[0.5,2]$, with stronger degradation only for very small or large thresholds.}
  \label{tab:tau_sensitivity}
\end{table*}

\begin{table*}[t]
  \centering
  \scriptsize
  \setlength{\tabcolsep}{5pt}
  \begin{tabular}{llcccccc}
    \toprule
    Condition & Token Budget ($k$)
      & No RRS
      & $[0,1]$
      & $[0.05,0.95]$
      & $[0.10,0.95]$
      & $[0.20,0.90]$
      & $[0.30,0.80]$ \\
    \midrule
    \multirow{4}{*}{$k=16$}
      & Clean
      & 0.268 & 0.252 & \textbf{0.275}
      & 0.272 & 0.266 & 0.260 \\
      & Gaussian ($\sigma=0.5$)
      & 0.238 & 0.232 & \textbf{0.249}
      & 0.246 & 0.246 & 0.230 \\
      & Gaussian ($\sigma=2.0$)
      & \textbf{0.174} & 0.151 & 0.171
      & 0.160 & 0.171 & \textbf{0.174} \\
      & S\&P ($\rho=0.10$)
      & 0.240 & 0.244 & \textbf{0.254}
      & 0.251 & 0.250 & 0.240 \\
    \midrule
    \multirow{4}{*}{$k=50$}
      & Clean
      & 0.306 & 0.281 & \textbf{0.308}
      & 0.306 & 0.307 & 0.298 \\
      & Gaussian ($\sigma=0.5$)
      & 0.265 & 0.259 & 0.272
      & \textbf{0.274} & 0.272 & 0.263 \\
      & Gaussian ($\sigma=2.0$)
      & \textbf{0.207} & 0.171 & 0.198
      & 0.197 & 0.203 & 0.204 \\
      & S\&P ($\rho=0.10$)
      & 0.284 & 0.272 & 0.294
      & \textbf{0.295} & 0.290 & 0.285 \\
    \midrule
    \multirow{4}{*}{$k=144$}
      & Clean
      & \textbf{0.323} & 0.301 & 0.310 & 0.315 & 0.317 & 0.318 \\
      & Gaussian ($\sigma=0.5$)
      & 0.317 & 0.301 & 0.324 & \textbf{0.329} & 0.311 & 0.306 \\
      & Gaussian ($\sigma=2.0$)
      & 0.185 & 0.168 & 0.183 & 0.192 & 0.190 & \textbf{0.204} \\
      & S\&P ($\rho=0.10$)
      & 0.315 & 0.307 & 0.321 & \textbf{0.320} & 0.319 & 0.301 \\
    \midrule
    \multicolumn{2}{l}{Mean}
& 0.260 & 0.245 & \textbf{0.263}
& \textbf{0.263} & 0.262 & 0.257 \\
    \bottomrule
  \end{tabular}
  \caption{Sensitivity analysis of the clipping interval applied
  to the RRS shrinkage coefficient $\alpha_i$ on MM-VET with
  LLaVA-1.5-7B. ``No RRS'' uses C+H token selection without residual
  shrinkage. The default clipping interval is $[0.10,0.95]$. Clipping is essential for preventing unstable shrinkage, whereas performance is relatively insensitive to the precise clipped interval. RRS improves performance under mild and sparse corruption but can be detrimental under severe dense Gaussian noise, where retaining the original selected tokens is preferable.}
  \label{tab:alpha_sensitivity}
\end{table*}




  
  
    
  

\clearpage
\bibliographystyle{unsrt}
\bibliography{mybib}




\end{document}